\documentclass[sigplan,10pt]{acmart}

\usepackage{amsmath}
\usepackage{enumitem}
\usepackage{xspace}
\usepackage{caption}
\usepackage{subcaption}
\usepackage{float}
\usepackage{cleveref}
\usepackage[small, compact]{titlesec}
\newcommand{\tech}[0]{PagedAttention\xspace}
\newcommand{\sys}[0]{vLLM\xspace}

\newcommand{\heading}[1]{\vspace{4pt}\noindent\textbf{#1}}

\usepackage{enumitem}
\newenvironment{CompactItemize}
  {\begin{itemize}[noitemsep,topsep=0pt,leftmargin=*]}
  {\end{itemize}}

\renewcommand{\shortauthors}{Woosuk Kwon et al.}

\setcopyright{rightsretained}

\begin{document}

\title{Efficient Memory Management for Large Language Model Serving with \emph{PagedAttention}}

\author{Woosuk Kwon$^{\text{1}, *}$\enskip Zhuohan Li$^{\text{1}, *}$\enskip Siyuan Zhuang$^{\text{1}}$\enskip Ying Sheng$^{\text{1}, \text{2}}$\enskip Lianmin Zheng$^{\text{1}}$\enskip Cody Hao Yu$^{\text{3}}$\enskip \\ Joseph E. Gonzalez$^{\text{1}}$\enskip Hao Zhang$^{\text{4}}$\enskip Ion Stoica$^{\text{1}}$}
\affiliation{\vspace{1mm} $^{\text{1}}$UC Berkeley \enskip $^{\text{2}}$Stanford University \enskip $^{\text{3}}$Independent Researcher \enskip $^{\text{4}}$UC San Diego \country{}}

\begin{abstract}
High throughput serving of large language models (LLMs) requires batching sufficiently many requests at a time.
However, existing systems struggle because the key-value cache (KV cache) memory for each request is huge and grows and shrinks dynamically.
When managed inefficiently, this memory can be significantly wasted by fragmentation and redundant duplication, limiting the batch size.
To address this problem, we propose \tech, an attention algorithm inspired by the classical virtual memory and paging techniques in operating systems.
On top of it, we build \sys, an LLM serving system that achieves (1) near-zero waste in KV cache memory and (2) flexible sharing of KV cache within and across requests to further reduce memory usage.
Our evaluations show that \sys improves the throughput of popular LLMs by 2-4$\times$ with the same level of latency compared to the state-of-the-art systems, such as FasterTransformer and Orca.
The improvement is more pronounced with longer sequences, larger models, and more complex decoding algorithms.
\sys's source code is publicly available at~\url{https://github.com/vllm-project/vllm}.
\end{abstract}

\acmYear{2023}\copyrightyear{2023}
\acmConference[SOSP '23]{ACM SIGOPS 29th Symposium on Operating Systems Principles}{October 23--26, 2023}{Koblenz, Germany}
\acmBooktitle{ACM SIGOPS 29th Symposium on Operating Systems Principles (SOSP '23), October 23--26, 2023, Koblenz, Germany}
\acmPrice{}
\acmDOI{10.1145/3600006.3613165}
\acmISBN{979-8-4007-0229-7/23/10}

\maketitle
\pagestyle{plain}

\section{Introduction}
\label{sec:intro}

The emergence of large language models (\emph{LLMs}) like GPT~\cite{openai2023gpt4,brown2020language} and PaLM~\cite{chowdhery2022palm} have enabled new applications such as programming assistants~\cite{copilot, chen2021evaluating} and universal chatbots~\cite{chatgpt, bard} that are starting to profoundly impact our work and daily routines. Many cloud companies~\cite{openaiapi, amazonbedrock} are racing to provide these applications as hosted services. However, running these applications is very expensive, requiring a large number of hardware accelerators such as GPUs. According to recent estimates, processing an LLM request can be 10$\times$ more expensive than a traditional keyword query~\cite{chat-cost}.
Given these high costs, increasing the throughput---and hence reducing the cost per request---of \emph{LLM serving} systems is becoming more important.

{\let\thefootnote\relax\footnote{{$^*$Equal contribution.}}\addtocounter{footnote}{-1}} At the core of LLMs lies an autoregressive Transformer model~\cite{vaswani2017attention}. 
This model generates words (tokens), \emph{one at a time}, based on the input (prompt) and the previous sequence of the output's tokens it has generated so far.
For each request, this expensive process is repeated until the model outputs a termination token.
This sequential generation process makes the workload \emph{memory-bound}, underutilizing the computation power of GPUs and limiting the serving throughput.

\begin{figure}
    \centering
    \begin{subfigure}[c]{0.40\columnwidth}
    \centering
    \includegraphics[width=\columnwidth]{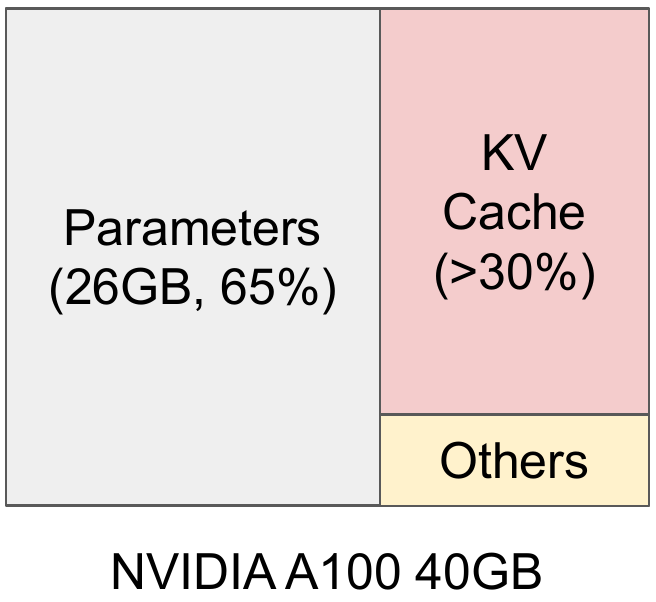}  
    \end{subfigure}
    \hfil
    \begin{subfigure}[c]{0.50\columnwidth}
    \centering
    \includegraphics[width=\columnwidth]{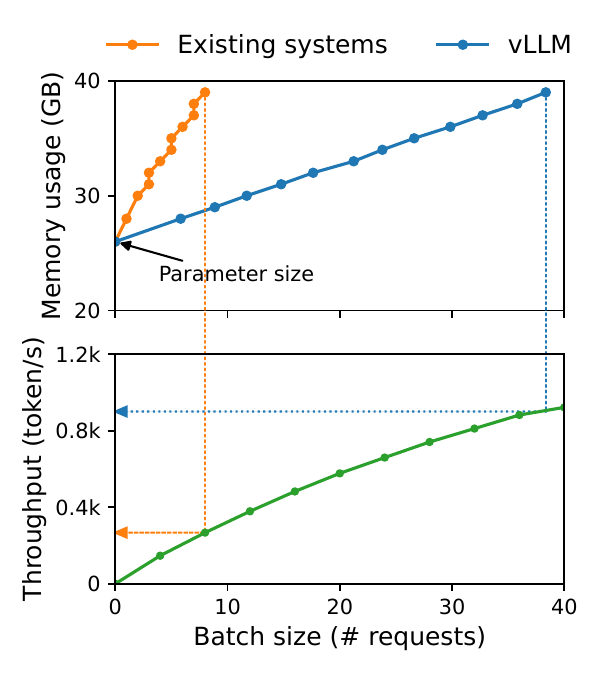}
    \end{subfigure}
    \vspace{-10pt}
    \caption{\emph{Left:} Memory layout when serving an LLM with 13B parameters on NVIDIA A100. The parameters (gray) persist in GPU memory throughout serving. The memory for the KV cache (red) is (de)allocated per serving request. A small amount of memory (yellow) is used ephemerally for activation. 
    \emph{Right:}
    \sys smooths out the rapid growth curve of KV cache memory seen in existing systems~\cite{nvidiaft,yu2022orca}, leading to a notable boost in serving throughput.}
    \label{fig:motivation}
\end{figure}

Improving the throughput is possible by batching multiple requests together.
However, to process many requests in a batch, the memory space for each request should be efficiently managed.
For example, Fig.~\ref{fig:motivation} (left) illustrates the memory distribution for a 13B-parameter LLM on an NVIDIA A100 GPU with 40GB RAM.
Approximately 65\% of the memory is allocated for the model weights, which remain static during serving.
Close to 30\% of the memory is used to store the dynamic states of the requests.
For Transformers, these states consist of the key and value tensors associated with the attention mechanism, commonly referred to as \emph{KV cache}~\cite{pope2022efficiently}, which represent the context from earlier tokens to generate new output tokens in sequence.
The remaining small percentage of memory is used for other data, including activations – the ephemeral tensors created when evaluating the LLM.
Since the model weights are constant and the activations only occupy a small fraction of the GPU memory, the way the KV cache is managed is critical in determining the maximum batch size.
When managed inefficiently, the KV cache memory can significantly limit the batch size and consequently the throughput of the LLM, as illustrated in Fig.~\ref{fig:motivation} (right).

In this paper, we observe that existing LLM serving systems~\cite{yu2022orca,nvidiaft} fall short of managing the KV cache memory efficiently.
This is mainly because they store the KV cache of a request in contiguous memory space, as most deep learning frameworks~\cite{paszke2019pytorch, olston2017tensorflow} require tensors to be stored in contiguous memory.
However, unlike the tensors in the traditional deep learning workloads, the KV cache has unique characteristics: it dynamically grows and shrinks over time as the model generates new tokens, and its lifetime and length are not known a priori.
These characteristics make the existing systems' approach significantly inefficient in two ways:

\begin{figure}[t]
    \centering
    \includegraphics[width=\columnwidth]{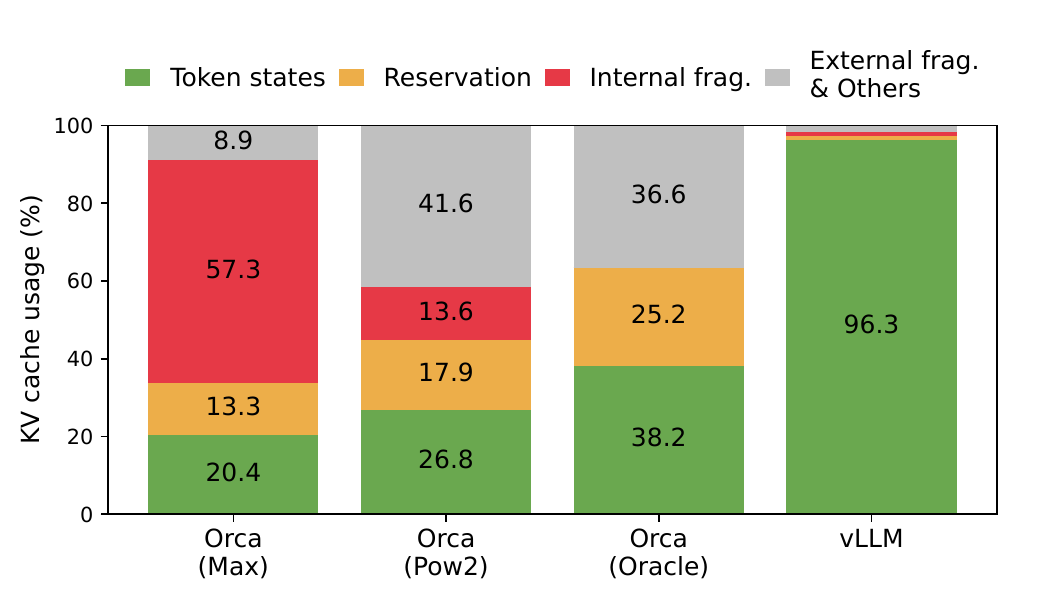}
    \caption{Average percentage of memory wastes in different LLM serving systems during the experiment in \S\ref{sec:eval:basic-sampling}.}
    \label{fig:memory-waste-percentage}
\end{figure}

First, the existing systems~\cite{yu2022orca,nvidiaft} suffer from internal and external memory fragmentation.
To store the KV cache of a request in contiguous space, they \emph{pre-allocate} a contiguous chunk of memory with the request's maximum length (e.g., 2048 tokens).
This can result in severe internal fragmentation, since the request's actual length can be much shorter than its maximum length (e.g., Fig.~\ref{fig:dataset-length-dist}).
Moreover, even if the actual length is known a priori, the pre-allocation is still inefficient: As the entire chunk is reserved during the request's lifetime, other shorter requests cannot utilize any part of the chunk that is currently unused.
Besides, external memory fragmentation can also be significant, since the pre-allocated size can be different for each request.
Indeed, our profiling results in Fig.~\ref{fig:memory-waste-percentage} show that only 20.4\% - 38.2\% of the KV cache memory is used to store the actual token states in the existing systems.

Second, the existing systems cannot exploit the opportunities for memory sharing.
LLM services often use advanced decoding algorithms, such as parallel sampling and beam search, that generate multiple outputs per request.
In these scenarios, the request consists of multiple sequences that can partially share their KV cache.
However, memory sharing is not possible in the existing systems because the KV cache of the sequences is stored in separate contiguous spaces.

To address the above limitations, we propose \emph{\tech}, an attention algorithm inspired by the operating system's (OS) solution to memory fragmentation and sharing: \emph{virtual memory with paging}.
\tech divides the request's KV cache into blocks, each of which can contain the attention keys and values of a fixed number of tokens.
In \tech, the blocks for the KV cache are not necessarily stored in contiguous space.
Therefore, we can manage the KV cache in a more flexible way as in OS’s virtual memory: one can think of blocks as pages, tokens as bytes, and requests as processes.
This design alleviates internal fragmentation by using relatively small blocks and allocating them on demand.
Moreover, it eliminates external fragmentation as all blocks have the same size.
Finally, it enables memory sharing at the granularity of a block, across the different sequences associated with the same request or even across the different requests.

In this work, we build \emph{\sys}, a high-throughput distributed LLM serving engine on top of \tech that achieves near-zero waste in KV cache memory.
vLLM uses block-level memory management and preemptive request scheduling that are co-designed with \tech.
\sys supports popular LLMs such as GPT~\cite{brown2020language}, OPT~\cite{zhang2022opt}, and LLaMA~\cite{touvron2023llama} with varying sizes, including the ones exceeding the memory capacity of a single GPU.
Our evaluations on various models and workloads show that \sys improves the LLM serving throughput by 2-4$\times$ compared to the state-of-the-art systems~\cite{yu2022orca,nvidiaft}, without affecting the model accuracy at all.
The improvements are more pronounced with longer sequences, larger models, and more complex decoding algorithms (\S\ref{sec:one-sequence-decoding-example}).
In summary, we make the following contributions:
\begin{CompactItemize}
    \item We identify the challenges in memory allocation in serving LLMs and quantify their impact on serving performance.
    \item We propose \tech, an attention algorithm that operates on KV cache stored in non-contiguous paged memory, which is inspired by the virtual memory and paging in OS.
    \item We design and implement \sys, a distributed LLM serving engine built on top of \tech.
    \item We evaluate \sys on various scenarios and demonstrate that it substantially outperforms the previous state-of-the-art solutions such as FasterTransformer~\cite{nvidiaft} and Orca~\cite{yu2022orca}.
\end{CompactItemize}

\section{Background}

In this section, we describe the generation and serving procedures of typical LLMs and the iteration-level scheduling used in LLM serving.

\subsection{Transformer-Based Large Language Models}
The task of language modeling is to model the probability of a list of tokens $(x_1, \ldots, x_n).$ Since language has a natural sequential ordering, it is common to factorize the joint probability over the whole sequence as the product of conditional probabilities (a.k.a. \emph{autoregressive decomposition} \cite{bengio2000neural}):
\begin{equation}
P(x) = P(x_1) \cdot P(x_2\mid x_1) \cdots P(x_n \mid x_1, \ldots, x_{n-1}). \label{eq:autoregressive-decomposition}   
\end{equation}

Transformers \cite{vaswani2017attention} have become the de facto standard architecture for modeling the probability above at a large scale. The most important component of a Transformer-based language model is its \emph{self-attention} layers. For an input hidden state sequence $(x_1, \ldots, x_n) \in \mathbb{R}^{n\times d}$, a self-attention layer first applies linear transformations on each position $i$ to get the query, key, and value vectors:
\begin{equation}
q_i = W_q x_i, \  k_i = W_k x_i, \  v_i = W_v x_i. \label{eq:qkv-transformation}
\end{equation}
Then, the self-attention layer computes the attention score $a_{ij}$ by multiplying the query vector at one position with all the key vectors before it and compute the output $o_i$ as the weighted average over the value vectors:
\begin{equation}
a_{ij} = \frac{\exp(q_i^\top k_j / \sqrt{d})}{\sum_{t=1}^{i}\exp(q_i^\top k_t / \sqrt{d})}, \ o_i = \sum_{j=1}^{i} a_{ij} v_j. \label{eq:attention-layer}
\end{equation}

Besides the computation in Eq.~\ref{eq:attention-layer}, all other components in the Transformer model, including the embedding layer, feed-forward layer, layer normalization \cite{ba2016layer}, residual connection \cite{he2016deep}, output logit computation, and the query, key, and value transformation in Eq.~\ref{eq:qkv-transformation}, are all applied independently position-wise in a form of
$y_i = f(x_i).$

\subsection{LLM Service \& Autoregressive Generation}

Once trained, LLMs are often deployed as a conditional generation service (e.g., completion API \cite{openaiapi} or chatbot \cite{chatgpt, bard}). A request to an LLM service provides a list of \emph{input prompt} tokens $(x_1, \ldots, x_n),$ and the LLM service generates a list of output tokens $(x_{n+1}, \ldots, x_{n+T})$ according to Eq.~\ref{eq:autoregressive-decomposition}. We refer to the concatenation of the prompt and output lists as \emph{sequence}. 

Due to the decomposition in Eq.~\ref{eq:autoregressive-decomposition}, the LLM can only sample and generate new tokens one by one, and the generation process of each new token depends on all the \emph{previous tokens} in that sequence, specifically their key and value vectors. In this sequential generation process, the key and value vectors of existing tokens are often cached for generating future tokens, known as \emph{KV cache}. Note that the KV cache of one token depends on all its previous tokens. This means that the KV cache of the same token appearing at different positions in a sequence will be different. 

Given a request prompt, the generation computation in the LLM service can be decomposed into two phases:

\heading{The prompt phase} takes the whole user prompt $(x_1, \ldots, x_n)$ as input and computes the probability of the first new token $P(x_{n+1} \mid x_1, \ldots, x_{n})$. During this process, also generates the key vectors $k_1, \ldots, k_n$ and value vectors $v_1, \ldots, v_n$.
Since prompt tokens $x_1, \ldots, x_n$ are all known, the computation of the prompt phase can be parallelized using matrix-matrix multiplication operations.
Therefore, this phase can efficiently use the parallelism inherent in GPUs.

\heading{The autoregressive generation phase} generates the remaining new tokens sequentially. 
At iteration $t$, the model takes one token $x_{n+t}$ as input and computes the probability $P(x_{n+t+1} \mid x_1, \ldots, x_{n+t})$ with the key vectors $k_1, \ldots, k_{n+t}$ and value vectors $v_1, \ldots, v_{n+t}$. Note that the key and value vectors at positions $1$ to $n + t - 1$ are cached at previous iterations, only the new key and value vector $k_{n+t}$ and $v_{n+t}$ are computed at this iteration. This phase completes either when the sequence reaches a maximum length (specified by users or limited by LLMs) or when an end-of-sequence (\emph{<eos>}) token is emitted.
The computation at different iterations cannot be parallelized due to the data dependency and often uses matrix-vector multiplication, which is less efficient. 
As a result, this phase severely underutilizes GPU computation and becomes memory-bound, being responsible for most portion of the latency of a single request.

\begin{figure*}[t]
    \centering
    \includegraphics[width=.85\linewidth]{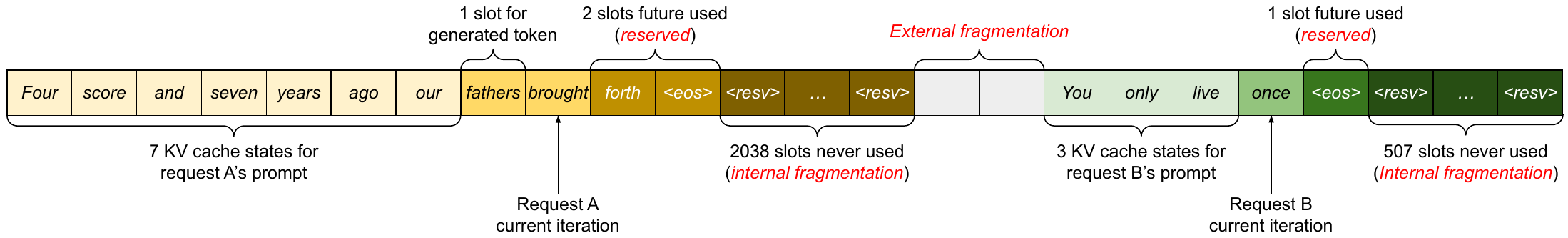}
    \vspace{-5pt}
    \caption{KV cache memory management in existing systems. Three types of memory wastes -- reserved, internal fragmentation, and external fragmentation -- exist that prevent other requests from fitting into the memory. The token in each memory slot represents its KV cache. Note the same tokens can have different KV cache when at different positions.}
    \label{fig:existing-memory-management}
\end{figure*}

\vspace{-1pt}
\subsection{Batching Techniques for LLMs}
\label{sec:iteration-level-scheduling}

The compute utilization in serving LLMs can be improved by batching multiple requests.
Because the requests share the same model weights, the overhead of moving weights is amortized across the requests in a batch, and can be overwhelmed by the computational overhead when the batch size is sufficiently large.
However, batching the requests to an LLM service is non-trivial for two reasons.
First, the requests may arrive at different times.
A naive batching strategy would either make earlier requests wait for later ones or delay the incoming requests until earlier ones finish, leading to significant queueing delays.
Second, the requests may have vastly different input and output lengths (Fig.~\ref{fig:dataset-length-dist}).
A straightforward batching technique would pad the inputs and outputs of the requests to equalize their lengths, wasting GPU computation and memory.

To address this problem, fine-grained batching mechanisms, such as cellular batching~\cite{gao2018low} and iteration-level scheduling~\cite{yu2022orca}, have been proposed.
Unlike traditional methods that work at the request level, these techniques operate at the iteration level.
After each iteration, completed requests are removed from the batch, and new ones are added.
Therefore, a new request can be processed after waiting for a single iteration, not waiting for the entire batch to complete.
Moreover, with special GPU kernels, these techniques eliminate the need to pad the inputs and outputs.
By reducing the queueing delay and the inefficiencies from padding, the fine-grained batching mechanisms significantly increase the throughput of LLM serving. 

\section{Memory Challenges in LLM Serving}

\label{sec:motivation}

Although fine-grained batching reduces the waste of computing and enables requests to be batched in a more flexible way,
the number of requests that can be batched together is still constrained by GPU memory capacity, particularly the space allocated to store the KV cache. 
In other words, the serving system's throughput is \emph{memory-bound}.
Overcoming this memory-bound requires addressing the following challenges in the memory management:

\heading{Large KV cache.} The KV Cache size grows quickly with the number of requests. As an example, for the 13B parameter OPT model \cite{zhang2022opt}, the KV cache of a single token demands 800 KB of space, calculated as $2$ (key and value vectors) $\times\  5120$ (hidden state size) $\times \  40$ (number of layers) $\times \  2$ (bytes per FP16). Since OPT can generate sequences up to 2048 tokens, the memory required to store the KV cache of one request can be as much as 1.6\,GB. Concurrent GPUs have memory capacities in the tens of GBs. Even if all available memory was allocated to KV cache, only a few tens of requests could be accommodated. Moreover, inefficient memory management can further decrease the batch size, as shown in Fig.~\ref{fig:memory-waste-percentage}. 
Additionally, given the current trends, the GPU's computation speed grows faster than the memory capacity \cite{gholami2021ai}. For example, from NVIDIA A100 to H100, The FLOPS increases by more than 2x, but the GPU memory stays at 80GB maximum. Therefore, we believe the memory will become an increasingly significant bottleneck.

\heading{Complex decoding algorithms.} 
LLM services offer a range of decoding algorithms for users to select from, each with varying implications for memory management complexity. For example, when users request multiple random samples from a single input prompt, a typical use case in program suggestion~\cite{copilot}, the KV cache of the prompt part, which accounts for 12\% of the total KV cache memory in our experiment (\S\ref{sec:eval:beamsearch}), can be shared to minimize memory usage. On the other hand, the KV cache during the autoregressive generation phase should remain unshared due to the different sample results and their dependence on context and position.
The extent of KV cache sharing depends on the specific decoding algorithm employed. In more sophisticated algorithms like beam search~\cite{sutskever2014sequence}, different request beams can share larger portions (up to 55\% memory saving, see \S\ref{sec:eval:beamsearch}) of their KV cache, and the sharing pattern evolves as the decoding process advances. 

\heading{Scheduling for unknown input \& output lengths.} The requests to an LLM service exhibit variability in their input and output lengths. This requires the memory management system to accommodate a wide range of prompt lengths. In addition, as the output length of a request grows at decoding, the memory required for its KV cache also expands and may exhaust available memory for incoming requests or ongoing generation for existing prompts. The system needs to make scheduling decisions, such as deleting or swapping out the KV cache of some requests from GPU memory.

\subsection{Memory Management in Existing Systems}

\label{sec:memory-management-in-existing-systems}

Since most operators in current deep learning frameworks \cite{paszke2019pytorch, olston2017tensorflow} require tensors to be stored in contiguous memory, previous LLM serving systems~\cite{yu2022orca,nvidiaft} also store the KV cache of one request as a contiguous tensor across the different positions. Due to the unpredictable output lengths from the LLM, they statically allocate a chunk of memory for a request based on the request's maximum possible sequence length, irrespective of the actual input or eventual output length of the request. 

Fig.~\ref{fig:existing-memory-management} illustrates two requests: request A with 2048 maximum possible sequence length and request B with a maximum of 512. The chunk pre-allocation scheme in existing systems has three primary sources of memory wastes: \emph{reserved} slots for future tokens, \emph{internal fragmentation} due to over-provisioning for potential maximum sequence lengths, and \emph{external fragmentation} from the memory allocator like the buddy allocator. The external fragmentation will never be used for generated tokens, which is known before serving a request. Internal fragmentation also remains unused, but this is only realized after a request has finished sampling. They are both pure memory waste. 
Although the reserved memory is eventually used, reserving this space for the entire request's duration, especially when the reserved space is large, occupies the space that could otherwise be used to process other requests. We visualize the average percentage of memory wastes in our experiments in Fig.~\ref{fig:memory-waste-percentage}, revealing that the actual effective memory in previous systems can be as low as 20.4\%.

Although compaction~\cite{wang2022pacman} has been proposed as a potential solution to fragmentation, performing compaction in a performance-sensitive LLM serving system is impractical due to the massive KV cache. Even with compaction, the pre-allocated chunk space for each request prevents memory sharing specific to decoding algorithms in existing memory management systems.

\section{Method}

\begin{figure}
    \centering
    \includegraphics[width=.9\columnwidth]{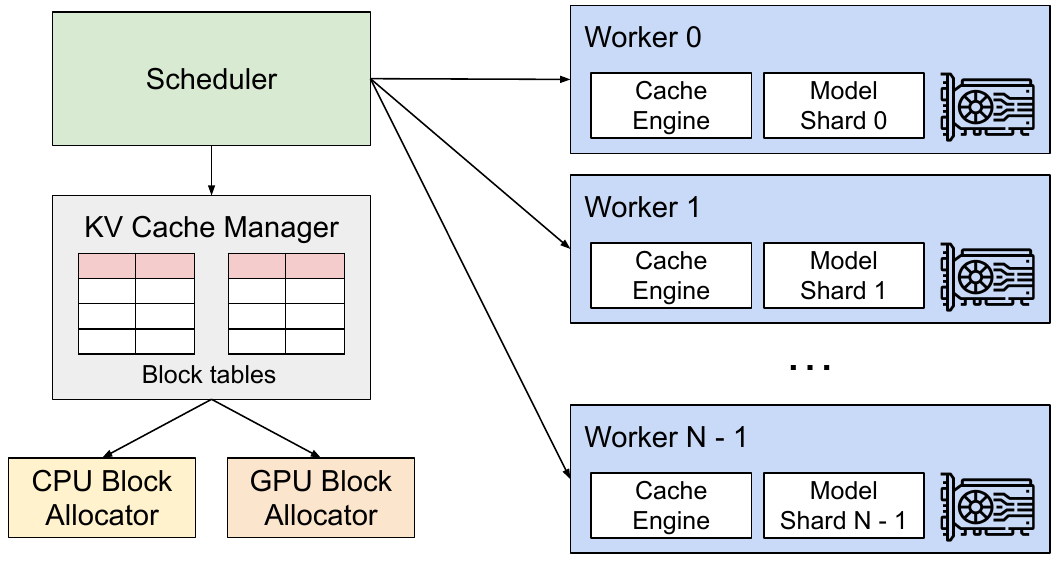}
    \vspace{-5pt}
    \caption{\sys system overview.} 
    \label{fig:system-overview}
    \vspace{-10pt}
\end{figure}

In this work, we develop a new attention algorithm, \emph{\tech}, and build an LLM serving engine, \emph{\sys}, to tackle the challenges outlined in \S\ref{sec:motivation}. The architecture of \sys is shown in Fig.~\ref{fig:system-overview}. \sys adopts a centralized scheduler to coordinate the execution of distributed GPU workers. The \emph{KV cache manager} effectively manages the KV cache in a paged fashion, enabled by \tech. Specifically, the KV cache manager manages the physical KV cache memory on the GPU workers through the instructions sent by the centralized scheduler.

Next, We describe the \tech algorithm in \S\ref{sec:pagedattention}. With that, we show the design of the KV cache manager in \S\ref{sec:block-space-manager} and how it facilitates \tech in \S\ref{sec:one-sequence-decoding-example}, respectively. Then, we show how this design facilitates effective memory management for various decoding methods (\S\ref{sec:decoding-scenerios}) and handles the variable length input and output sequences (\S\ref{sec:scheduling}). Finally, we show how the system design of \sys works in a distributed setting (\S\ref{sec:distributed}). 

\subsection{\tech}
\label{sec:pagedattention}

To address the memory challenges in \S\ref{sec:motivation}, we introduce \emph{\tech}, an attention algorithm inspired by the classic idea of \emph{paging} \cite{kilburn1962one} in operating systems. Unlike the traditional attention algorithms, \tech allows storing continuous keys and values in non-contiguous memory space. Specifically, \tech partitions the KV cache of each sequence into \emph{KV blocks}. Each block contains the key and value vectors for a fixed number of tokens,\footnote{In Transformer, each token has a set of key and value vectors across layers and attention heads within a layer. All the key and value vectors can be managed together within a single KV block, or the key and value vectors at different heads and layers can each have a separate block and be managed in separate block tables. The two designs have no performance difference and we choose the second one for easy implementation.}  which we denote as \emph{KV block size} ($B$). Denote the key block $K_j = (k_{(j - 1) B + 1}, \ldots, k_{jB})$ and value block $V_j = (v_{(j - 1) B + 1}, \ldots, v_{jB}).$ The attention computation in Eq.~\ref{eq:attention-layer} can be transformed into the following block-wise computation:
\begin{equation}
A_{ij} = \frac{\exp(q_i^\top K_j / \sqrt{d})}{\sum_{t=1}^{\lceil i/B \rceil}\exp(q_i^\top K_t\mathbf{1} / \sqrt{d})}, \ o_i = \sum_{j=1}^{\lceil i/B \rceil} V_j A_{ij}^\top, \label{eq:attention-layer}
\end{equation}
where $A_{ij} = (a_{i,(j - 1) B + 1}, \ldots, a_{i,jB})$ is the row vector of attention score on $j$-th KV block.

\begin{figure}
    \centering
    \includegraphics[width=.8\columnwidth]{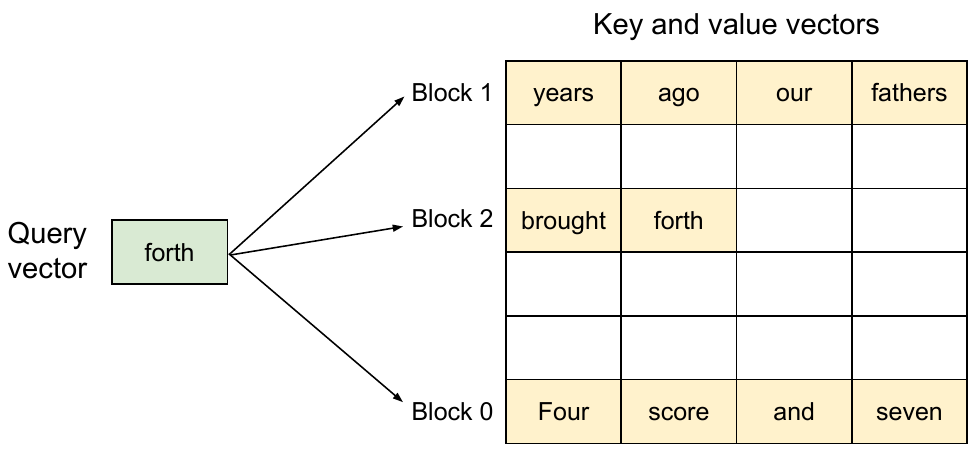}
    \vspace{-5pt}
    \caption{Illustration of the \tech algorithm, where the attention key and values vectors are stored as non-contiguous blocks in the memory. }
    \label{fig:pagedattention}
    \vspace{-10pt}
\end{figure}

During the attention computation, the \tech kernel identifies and fetches different KV blocks separately. We show an example of \tech in Fig.~\ref{fig:pagedattention}: The key and value vectors are spread across three blocks, and the three blocks are not contiguous on the physical memory. At each time, the kernel multiplies the query vector $q_i$ of the query token (``\emph{forth}'') and the key vectors $K_j$ in a block (e.g., key vectors of ``\emph{Four score and seven}'' for block 0) to compute the attention score $A_{ij},$ and later multiplies $A_{ij}$ with the value vectors $V_j$ in a block to derive the final attention output $o_i.$ 

In summary, the \tech algorithm allows the KV blocks to be stored in non-contiguous physical memory, which enables more flexible paged memory management in \sys.

\subsection{KV Cache Manager}
\label{sec:block-space-manager}

The key idea behind \sys's memory manager is analogous to the \emph{virtual memory} \cite{kilburn1962one} in operating systems. OS partitions memory into fixed-sized \emph{pages} and maps user programs' logical pages to physical pages. Contiguous logical pages can correspond to non-contiguous physical memory pages, allowing user programs to access memory as though it were contiguous. Moreover, physical memory space needs not to be fully reserved in advance, enabling the OS to dynamically allocate physical pages as needed. \sys uses the ideas behind virtual memory to manage the KV cache in an LLM service. Enabled by \tech, we organize the KV cache as fixed-size KV blocks, like pages in virtual memory. 

A request's KV cache is represented as a series of \emph{logical KV blocks}, filled from left to right as new tokens and their KV cache are generated. 
The last KV block's unfilled positions are reserved for future generations.
On GPU workers, a \emph{block engine} allocates a contiguous chunk of GPU DRAM and divides it into \emph{physical KV blocks} (this is also done on CPU RAM for swapping; see \S\ref{sec:scheduling}). The \emph{KV block manager} also maintains \emph{block tables}---the mapping between logical and physical KV blocks of each request. Each block table entry records the corresponding physical blocks of a logical block and the number of filled positions. Separating logical and physical KV blocks allows \sys to dynamically grow the KV cache memory without reserving it for all positions in advance, which eliminates most memory waste in existing systems, as in Fig.~\ref{fig:memory-waste-percentage}. 

\subsection{Decoding with \tech and \sys}
\label{sec:one-sequence-decoding-example}

\begin{figure}
    \centering
    \includegraphics[width=\columnwidth]{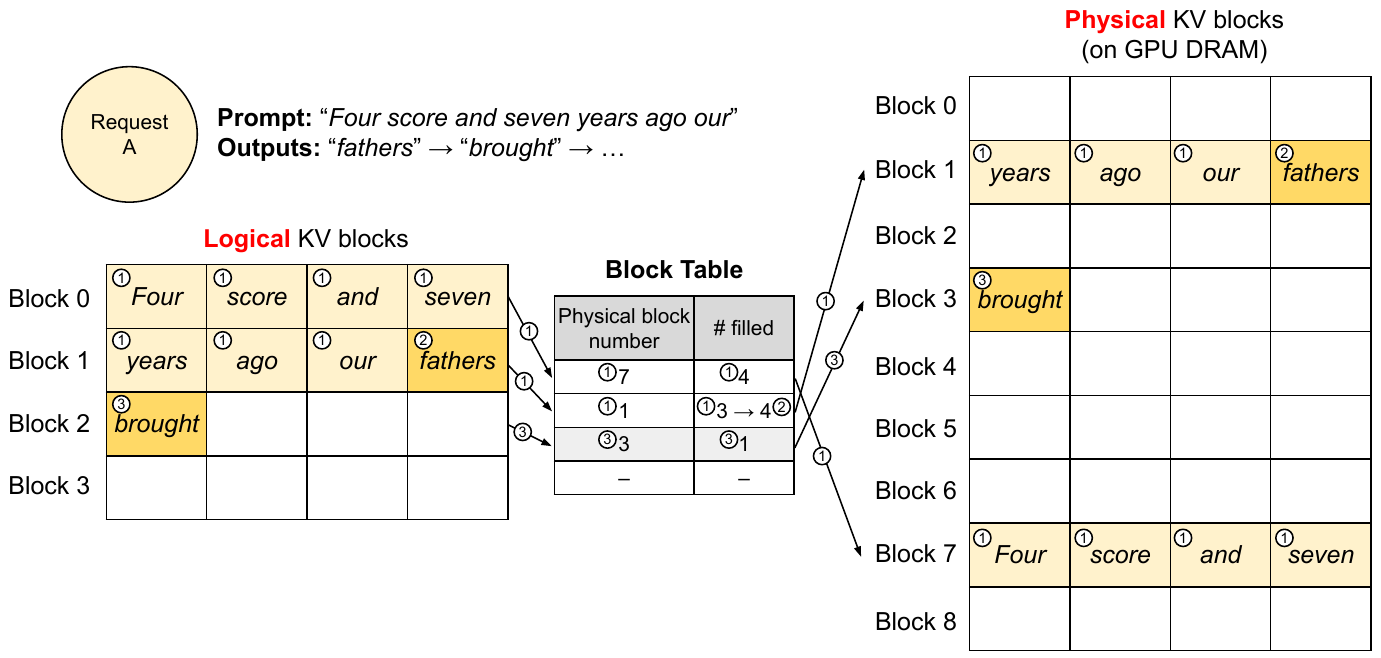}
    \vspace{-15pt}
    \caption{Block table translation in \sys. }
    \label{fig:block-table-translation}
    \vspace{-15pt}
\end{figure}

Next, we walk through an example, as in Fig.~\ref{fig:block-table-translation}, to demonstrate how \sys executes \tech and manages the memory during the decoding process of a single input sequence: \textcircled{\small 1} As in OS's virtual memory, \sys does not require reserving the memory for the maximum possible generated sequence length initially.
Instead, it reserves only the necessary KV blocks to accommodate the KV cache generated during prompt computation. In this case, The prompt has 7 tokens, so \sys maps the first 2 logical KV blocks (0 and 1) to 2 physical KV blocks (7 and 1, respectively). In the prefill step, \sys generates the KV cache of the prompts and the first output token with a conventional self-attention algorithm (e.g., \cite{dao2022flashattention}). \sys then stores the KV cache of the first 4 tokens in logical block 0 and the following 3 tokens in logical block 1. The remaining slot is reserved for the subsequent autoregressive generation phase. 
\textcircled{\small 2} In the first autoregressive decoding step, \sys generates the new token with the \tech algorithm on physical blocks 7 and 1. Since one slot remains available in the last logical block, the newly generated KV cache is stored there, and the block table's \#filled record is updated. \textcircled{\small 3} At the second decoding step, as the last logical block is full, \sys stores the newly generated KV cache in a new logical block; \sys allocates a new physical block (physical block 3) for it and stores this mapping in the block table.

Globally, for each decoding iteration, \sys first selects a set of candidate sequences for batching (more in \S\ref{sec:scheduling}), and allocates the physical blocks for the newly required logical blocks. Then, \sys concatenates all the input tokens of the current iteration (i.e., all tokens for prompt phase requests and the latest tokens for generation phase requests) as one sequence and feeds it into the LLM. During LLM's computation, \sys uses the \tech kernel to access the previous KV cache stored in the form of logical KV blocks and saves the newly generated KV cache into the physical KV blocks. 
Storing multiple tokens within a KV block (block size > 1) enables the \tech kernel to process the KV cache across more positions in parallel, thus increasing the hardware utilization and reducing latency. However, a larger block size also increases memory fragmentation. We study the effect of block size in \S\ref{sec:eval:blocksize}.

\begin{figure}
    \centering
    \includegraphics[width=\columnwidth]{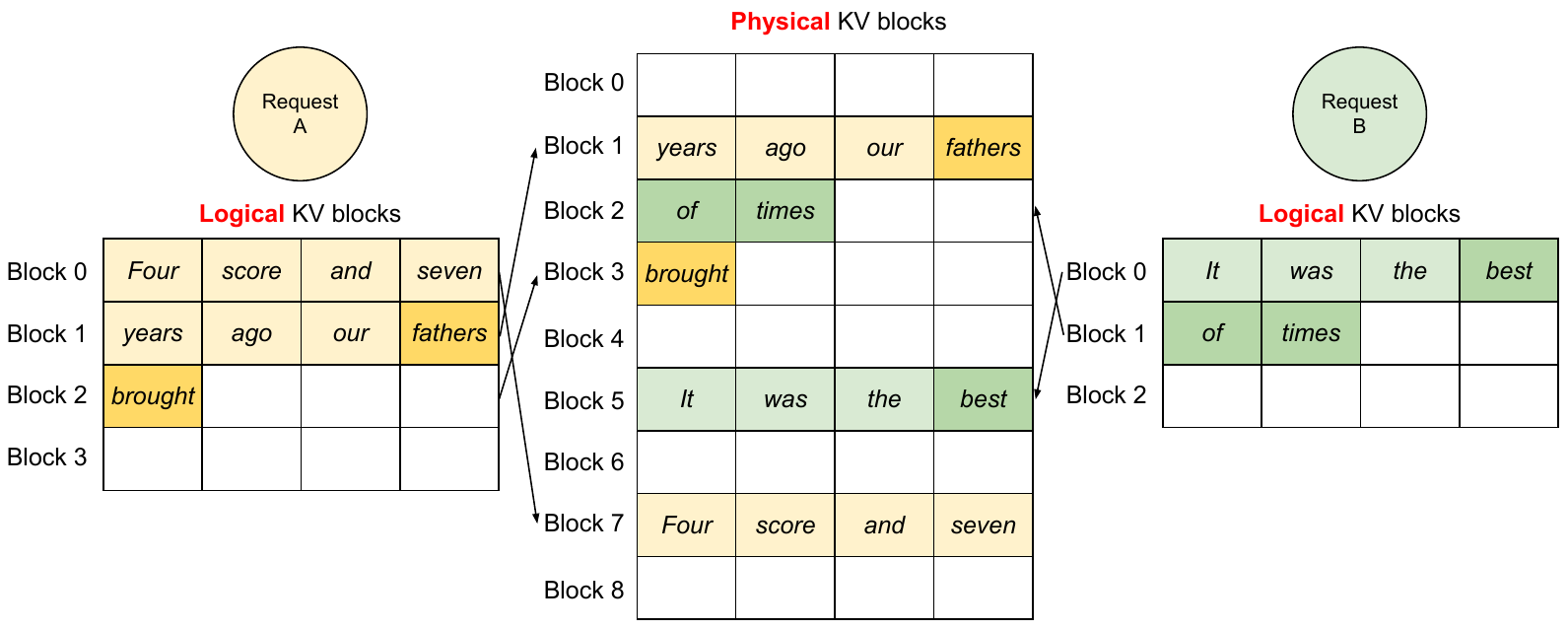}
    \vspace{-15pt}
    \caption{Storing the KV cache of two requests at the same time in \sys.}
    \label{fig:multi-sequence-blocks}
    \vspace{-10pt}
\end{figure}

Again, \sys dynamically assigns new physical blocks to logical blocks as more tokens and their KV cache are generated. As all the blocks are filled from left to right and a new physical block is only allocated when all previous blocks are full, \sys limits all the memory wastes for a request within one block, so it can effectively utilize all the memory, as shown in Fig.~\ref{fig:memory-waste-percentage}. This allows more requests to fit into memory for batching---hence improving the throughput. Once a request finishes its generation, its KV blocks can be freed to store the KV cache of other requests. 
In Fig.~\ref{fig:multi-sequence-blocks}, we show an example of \sys managing the memory for two sequences. The logical blocks of the two sequences are mapped to different physical blocks within the space reserved by the block engine in GPU workers. The neighboring logical blocks of both sequences do not need to be contiguous in physical GPU memory and the space of physical blocks can be effectively utilized by both sequences.

\subsection{Application to Other Decoding Scenarios}
\label{sec:decoding-scenerios}

\S\ref{sec:one-sequence-decoding-example} shows how \tech and \sys handle basic decoding algorithms, such as greedy decoding and sampling, that take one user prompt as input and generate a single output sequence. In many successful LLM applications~\cite{copilot, openaiapi}, an LLM service must offer more complex decoding scenarios that exhibit complex accessing patterns and more opportunities for memory sharing. We show the general applicability of \sys on them in this section. 

\heading{Parallel sampling.} 
In LLM-based program assistants~\cite{chen2021evaluating, copilot}, an LLM generates multiple sampled outputs for a single input prompt; users can choose a favorite output from various candidates. 
So far we have implicitly assumed that a request generates a single sequence. In the remainder of this paper, we assume the more general case in which a request generates multiple sequences.
In parallel sampling, one request includes multiple samples sharing the same input prompt, allowing the KV cache of the prompt to be shared as well. Via its \tech and paged memory management, \sys can realize this sharing easily and save memory. 

\begin{figure}
    \centering
    \includegraphics[width=\columnwidth]{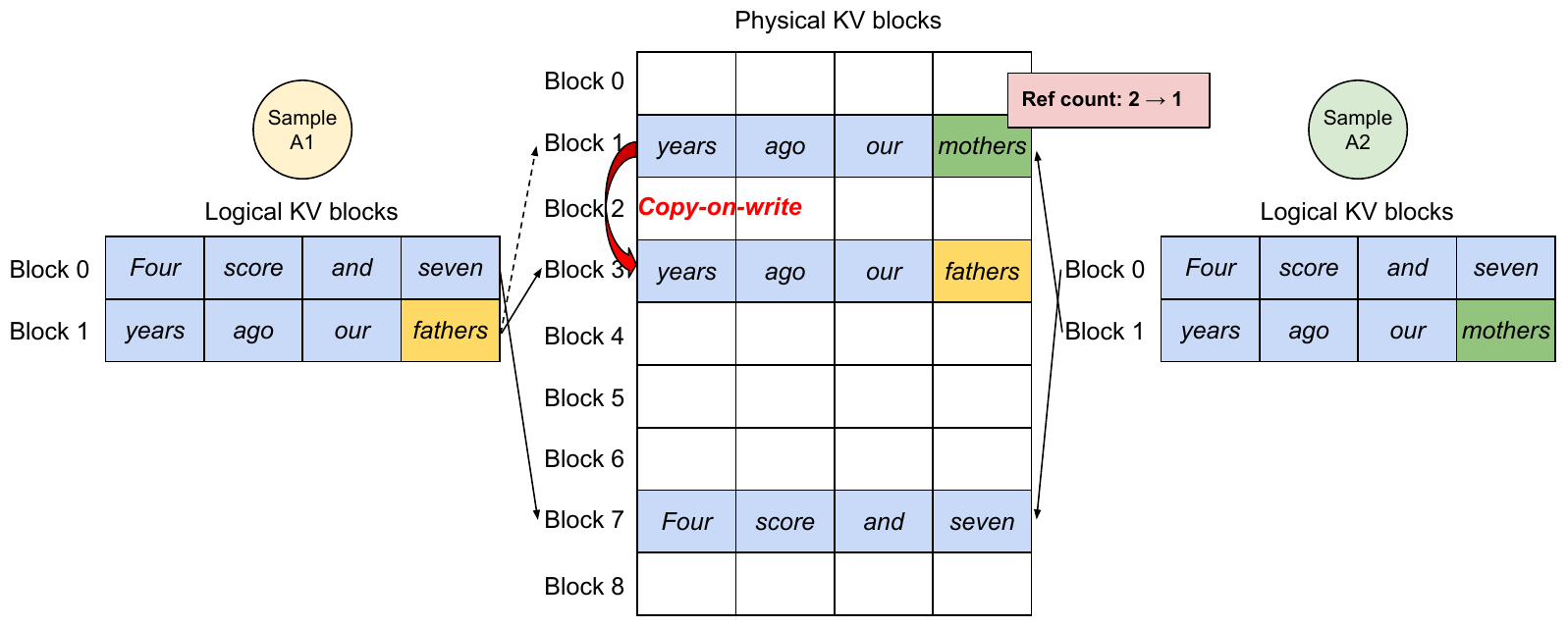}
    \vspace{-15pt}
    \caption{Parallel sampling example.}
    \label{fig:parallel-decoding}
    \vspace{-15pt}
\end{figure}

Fig.~\ref{fig:parallel-decoding} shows an example of parallel decoding for two outputs. 
Since both outputs share the same prompt, we only reserve space for one copy of the prompt's state at the prompt phase; the logical blocks for the prompts of both sequences are mapped to the same physical blocks: the logical block 0 and 1 of both sequences are mapped to physical blocks 7 and 1, respectively. 
Since a single physical block can be mapped to multiple logical blocks, we introduce a \emph{reference count} for each physical block. In this case, the reference counts for physical blocks 7 and 1 are both 2. At the generation phase, the two outputs sample different output tokens and need separate storage for KV cache. 
\sys implements a \emph{copy-on-write} mechanism at the block granularity for the physical blocks that need modification by multiple sequences, similar to the copy-on-write technique in OS virtual memory (e.g., when forking a process).
Specifically, in Fig.~\ref{fig:parallel-decoding}, when sample A1 needs to write to its last logical block (logical block 1), \sys recognizes that the reference count of the corresponding physical block (physical block 1) is greater than 1; it allocates a new physical block (physical block 3), instructs the block engine to copy the information from physical block 1, and decreases the reference count to 1. Next, when sample A2 writes to physical block 1, the reference count is already reduced to 1; thus A2 directly writes its newly generated KV cache to physical block 1. 

In summary, \sys enables the sharing of most of the space used to store the prompts' KV cache across multiple output samples, with the exception of the final logical block, which is managed by a copy-on-write mechanism. By sharing physical blocks across multiple samples, memory usage can be greatly reduced, especially for \emph{long input prompts}. 

\heading{Beam search.} In LLM tasks like machine translation~\cite{wu2016google}, the users expect the top-$k$ most appropriate translations output by the LLM. Beam search \cite{sutskever2014sequence} is widely used to decode the most probable output sequence from an LLM, as it mitigates the computational complexity of fully traversing the sample space.
The algorithm relies on the \emph{beam width} parameter $k$, which determines the number of top candidates retained at every step. During decoding, beam search expands each candidate sequence in the beam by considering all possible tokens, computes their respective probabilities using the LLM, and retains the top-$k$ most probable sequences out of $k \cdot |V|$ candidates, where $|V|$ is the vocabulary size.

\begin{figure}
    \centering
    \includegraphics[width=0.9\columnwidth]{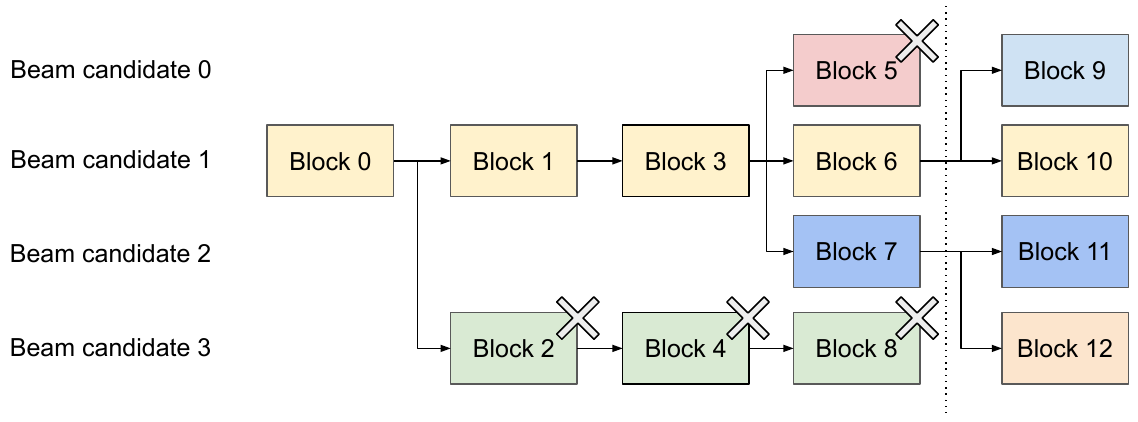}
    \vspace{-10pt}
    \caption{Beam search example.}
    \label{fig:beam-search}
    \vspace{-10pt}
\end{figure}

Unlike parallel decoding, beam search facilities sharing not only the initial prompt blocks but also other blocks across different candidates, and the sharing patterns dynamically change as the decoding process advances, similar to the process tree in the OS created by compound forks. 
Fig.~\ref{fig:beam-search} shows how \sys manages the KV blocks for a beam search example with $k = 4$. Prior to the iteration illustrated as the dotted line, each candidate sequence has used 4 full logical blocks. All beam candidates share the first block 0 (i.e., prompt). Candidate 3 digresses from others from the second block. Candidates 0-2 share the first 3 blocks and diverge at the fourth block. At subsequent iterations, the top-4 probable candidates all originate from candidates 1 and 2. As the original candidates 0 and 3 are no longer among the top candidates, their logical blocks are freed, and the reference counts of corresponding physical blocks are reduced. \sys frees all physical blocks whose reference counts reach 0 (blocks 2, 4, 5, 8). Then, \sys allocates new physical blocks (blocks 9-12) to store the new KV cache from the new candidates. Now, all candidates share blocks 0, 1, 3; candidates 0 and 1 share block 6, and candidates 2 and 3 further share block 7.

Previous LLM serving systems require frequent memory copies of the KV cache across the beam candidates. For example, in the case shown in Fig.~\ref{fig:beam-search}, after the dotted line, candidate 3 would need to copy a large portion of candidate 2's KV cache to continue generation. This frequent memory copy overhead is significantly reduced by \sys's physical block sharing. In \sys, most blocks of different beam candidates can be shared. The copy-on-write mechanism is applied only when the newly generated tokens are within an old shared block, as in parallel decoding.  This involves only copying one block of data.

\begin{figure}
    \centering
    \includegraphics[width=.9\columnwidth]{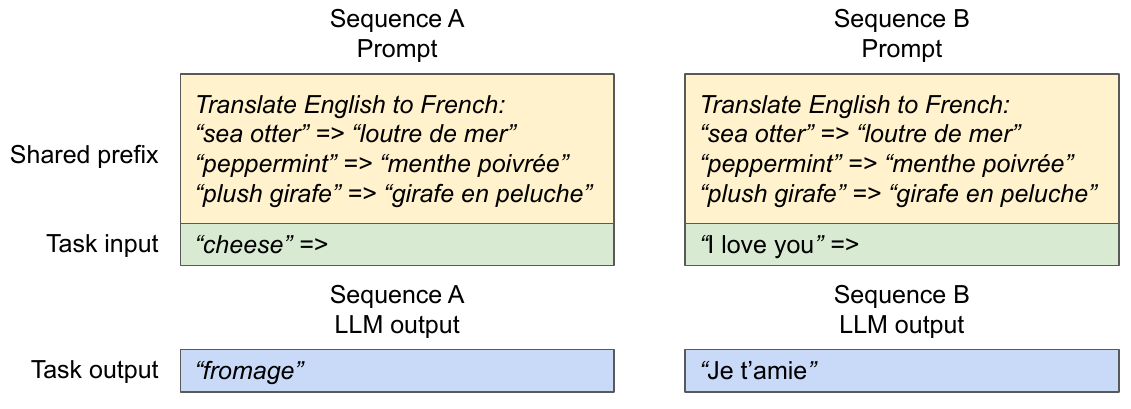}
    \vspace{-5pt}
    \caption{Shared prompt example for machine translation. The examples are adopted from~\cite{brown2020language}.}
    \label{fig:share-prompt}
    \vspace{-15pt}
\end{figure}

\heading{Shared prefix.} 
Commonly, the LLM user provides a (long) description of the task including instructions and example inputs and outputs, also known as \emph{system prompt} \cite{chatgptuserprompt}.
The description is concatenated with the actual task input to form the prompt of the request. The LLM generates outputs based on the full prompt. Fig.~\ref{fig:share-prompt} shows an example. Moreover, the shared prefix can be further tuned, via prompt engineering, to improve the accuracy of the downstream tasks \cite{li2021prefix, lester2021power}.

For this type of application, many user prompts share a prefix, thus the LLM service provider can store the KV cache of the prefix in advance to reduce the redundant computation spent on the prefix. In \sys, this can be conveniently achieved by reserving a set of physical blocks for a set of predefined shared prefixes by the LLM service provider, as how OS handles shared library across processes. A user input prompt with the shared prefix can simply map its logical blocks to the cached physical blocks (with the last block marked copy-on-write). The prompt phase computation only needs to execute on the user's task input.

\heading{Mixed decoding methods.} 
The decoding methods discussed earlier exhibit diverse memory sharing and accessing patterns. Nonetheless, \sys facilitates the simultaneous processing of requests with different decoding preferences, which existing systems \emph{cannot} efficiently do.
This is because \sys conceals the complex memory sharing between different sequences via a common mapping layer that translates logical blocks to physical blocks. The LLM and its execution kernel only see a list of physical block IDs for each sequence and do not need to handle sharing patterns across sequences. Compared to existing systems, this approach broadens the batching opportunities for requests with different sampling requirements, ultimately increasing the system's overall throughput.

\subsection{Scheduling and Preemption}
\label{sec:scheduling}

When the request traffic surpasses the system’s capacity, \sys must prioritize a subset of requests. In \sys, we adopt the first-come-first-serve (FCFS) scheduling
policy for all requests, ensuring fairness and preventing starvation. When \sys needs to preempt requests, it ensures
that the earliest arrived requests are served first and the
latest requests are preempted first.

LLM services face a unique challenge: the input prompts for an LLM can vary significantly in length, and the resulting output lengths are not known a priori, contingent on both the input prompt and the model. As the number of requests and their outputs grow, \sys can run out of the GPU's physical blocks to store the newly generated KV cache. There are two classic questions that \sys needs to answer in this context: (1) Which blocks should it evict? (2) How to recover evicted blocks if needed again? Typically, eviction policies use heuristics to predict which block will be accessed furthest in the future and evict that block. Since in our case we know that all blocks of a sequence are accessed together, we implement an all-or-nothing eviction policy, i.e., either evict all or none of the blocks of a sequence. Furthermore, multiple sequences within one request (e.g., beam candidates in one beam search request) are gang-scheduled as a \emph{sequence group}. The sequences within one sequence group are always preempted or rescheduled together due to potential memory sharing across those sequences. 
To answer the second question of how to recover an evicted block, we consider two techniques:

\heading{Swapping.} This is the classic technique used by most virtual memory implementations which copy the evicted pages to a swap space on the disk. In our case, we copy evicted blocks to the CPU memory. As shown in Fig.~\ref{fig:system-overview}, besides the GPU block allocator, \sys includes a CPU block allocator to manage the physical blocks swapped to CPU RAM. When \sys exhausts free physical blocks for new tokens, it selects a set of sequences to evict and transfer their KV cache to the CPU. Once it preempts a sequence and evicts its blocks, \sys stops accepting new requests until all preempted sequences are completed. Once a request completes, its blocks are freed from memory, and the blocks of a preempted sequence are brought back in to continue the processing of that sequence.
Note that with this design, the number of blocks swapped to the CPU RAM never exceeds the number of total physical blocks in the GPU RAM, so the swap space on the CPU RAM is bounded by the GPU memory allocated for the KV cache. 

\heading{Recomputation.} In this case, we simply recompute the KV cache when the preempted sequences are rescheduled. Note that recomputation latency can be significantly lower than the original latency, as the tokens generated at decoding can be concatenated with the original user prompt as a new prompt---their KV cache at all positions can be generated in one prompt phase iteration.

The performances of swapping and recomputation depend on the bandwidth between CPU RAM and GPU memory and the computation power of the GPU. We examine the speeds of swapping and recomputation in \S\ref{sec:eval:scheduling}.

\subsection{Distributed Execution}
\label{sec:distributed}
Many LLMs have parameter sizes exceeding the capacity of a single GPU~\cite{brown2020language, chowdhery2022palm}. Therefore, it is necessary to partition them across distributed GPUs and execute them in a model parallel fashion~\cite{zheng2022alpa, li2023alpaserve}. This calls for a memory manager capable of handling distributed memory. \sys is effective in distributed settings by supporting the widely used Megatron-LM style tensor model parallelism strategy on Transformers~\cite{shoeybi2019megatron}. 
This strategy adheres to an SPMD (Single Program Multiple Data) execution schedule, wherein the linear layers are partitioned to perform block-wise matrix multiplication, and the the GPUs constantly synchronize intermediate results via an all-reduce operation. Specifically, the attention operator is split on the attention head dimension, each SPMD process takes care of a subset of attention heads in multi-head attention.

We observe that even with model parallel execution, each model shard still processes the same set of input tokens, thus requiring the KV Cache for the same positions. Therefore, \sys features a single KV cache manager within the centralized scheduler, as in Fig.~\ref{fig:system-overview}. Different GPU workers share the manager, as well as the mapping from logical blocks to physical blocks. 
This common mapping allows GPU workers to execute the model with the physical blocks provided by the scheduler for each input request.
Although each GPU worker has the same physical block IDs, a worker only stores a portion of the KV cache for its corresponding attention heads.

In each step, the scheduler first prepares the message with input token IDs for each request in the batch, as well as the block table for each request. Next, the scheduler broadcasts this control message to the GPU workers. Then, the GPU workers start to execute the model with the input token IDs. In the attention layers, the GPU workers read the KV cache according to the block table in the control message. During execution, the GPU workers synchronize the intermediate results with the all-reduce communication primitive without the coordination of the scheduler, as in \cite{shoeybi2019megatron}. In the end, the GPU workers send the sampled tokens of this iteration back to the scheduler. In summary, GPU workers do not need to synchronize on memory management as they only need to receive all the memory management information at the beginning of each decoding iteration along with the step inputs.

\section{Implementation}
\label{sec:impl}

\begin{table}[tp]\centering\small
\caption{Model sizes and server configurations.}
\vspace{-10pt}
\scalebox{0.9}{
\begin{tabular}{@{} l c c c @{}} \toprule
Model size & \textbf{13B} & \textbf{66B} & \textbf{175B} \\
\midrule
GPUs & A100 & 4$\times$A100 & 8$\times$A100-80GB \\
Total GPU memory & 40 GB & 160 GB & 640 GB \\
Parameter size & 26 GB & 132 GB & 346 GB \\
\midrule
Memory for KV cache & 12 GB & 21 GB & 264 GB \\
Max. \# KV cache slots & 15.7K & 9.7K & 60.1K \\
\bottomrule
\end{tabular}
}
\vspace{-5pt}
\label{table:model_config}
\end{table}

\sys is an end-to-end serving system with a FastAPI~\cite{fastapi} frontend and a GPU-based inference engine.
The frontend extends the OpenAI API~\cite{openaiapi} interface, allowing users to customize sampling parameters for each request, such as the maximum sequence length and the beam width $k$.
The \sys engine is written in 8.5K lines of Python and 2K lines of C++/CUDA code.
We develop control-related components including the scheduler and the block manager in Python while developing custom CUDA kernels for key operations such as \tech.
For the model executor, we implement popular LLMs such as GPT~\cite{brown2020language}, OPT~\cite{zhang2022opt}, and LLaMA~\cite{touvron2023llama} using PyTorch~\cite{paszke2019pytorch} and Transformers~\cite{wolf2020transformers}.
We use NCCL~\cite{nccl} for tensor communication across the distributed GPU workers.

\subsection{Kernel-level Optimization}
\label{sec:impl-kernel}

Since \tech introduces memory access patterns that are not efficiently supported by existing systems, we develop several GPU kernels for optimizing it.
(1) \emph{Fused reshape and block write.}
In every Transformer layer, the new KV cache are split into blocks, reshaped to a memory layout optimized for block read, then saved at positions specified by the block table. To minimize kernel launch overheads, we fuse them into a single kernel.
(2) \emph{Fusing block read and attention.}
We adapt the attention kernel in FasterTransformer~\cite{nvidiaft} to read KV cache according to the block table and perform attention operations on the fly.
To ensure coalesced memory access, we assign a GPU warp to read each block.
Moreover, we add support for variable sequence lengths within a request batch.
(3) \emph{Fused block copy.}
Block copy operations, issued by the copy-on-write mechanism, may operate on discontinuous blocks.
This can lead to numerous invocations of small data movements if we use the \texttt{cudaMemcpyAsync} API.
To mitigate the overhead, we implement a kernel that batches the copy operations for different blocks into a single kernel launch.

\begin{figure}[t]
     \centering
     \begin{subfigure}[b]{0.5\linewidth}
         \centering
         \includegraphics[width=1\columnwidth]{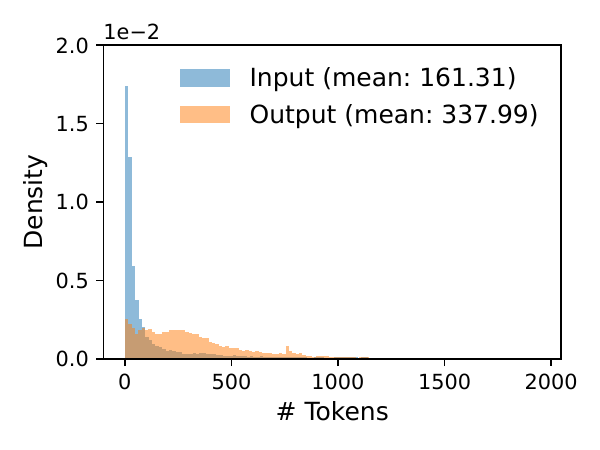}
         \vskip -0.1in
         \caption{\small ShareGPT}
     \end{subfigure}
     \begin{subfigure}[b]{0.5\linewidth}
         \centering
         \includegraphics[width=1\columnwidth]{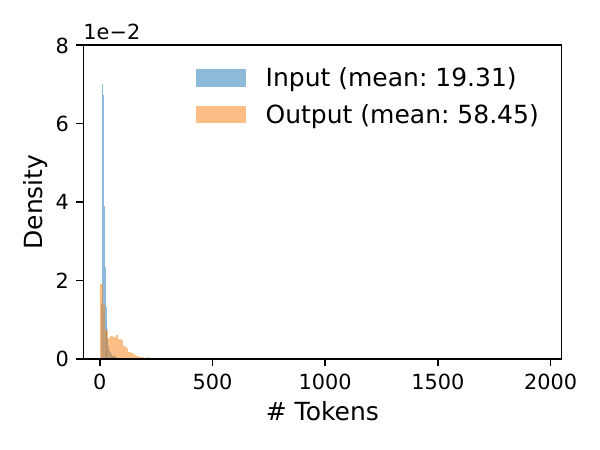}
         \vskip -0.1in
         \caption{\small Alpaca}
     \end{subfigure}
     \vskip -0.1in
     \caption{Input and output length distributions of the (a) ShareGPT and (b) Alpaca datasets.}     
     \vspace{-10pt}
\label{fig:dataset-length-dist}
\end{figure}

\subsection{Supporting Various Decoding Algorithms}

\sys implements various decoding algorithms using three key methods: \texttt{fork}, \texttt{append}, and \texttt{free}.
The \texttt{fork} method creates a new sequence from an existing one.
The \texttt{append} method appends a new token to the sequence.
Finally, the \texttt{free} method deletes the sequence.
For instance, in parallel sampling, \sys creates multiple output sequences from the single input sequence using the \texttt{fork} method.
It then adds new tokens to these sequences in every iteration with \texttt{append}, and deletes sequences that meet a stopping condition using \texttt{free}.
The same strategy is also applied in beam search and prefix sharing by \sys.
We believe future decoding algorithms can also be supported by combining these methods.

\begin{figure*}[t]
    \centering
    \includegraphics[scale=0.46]{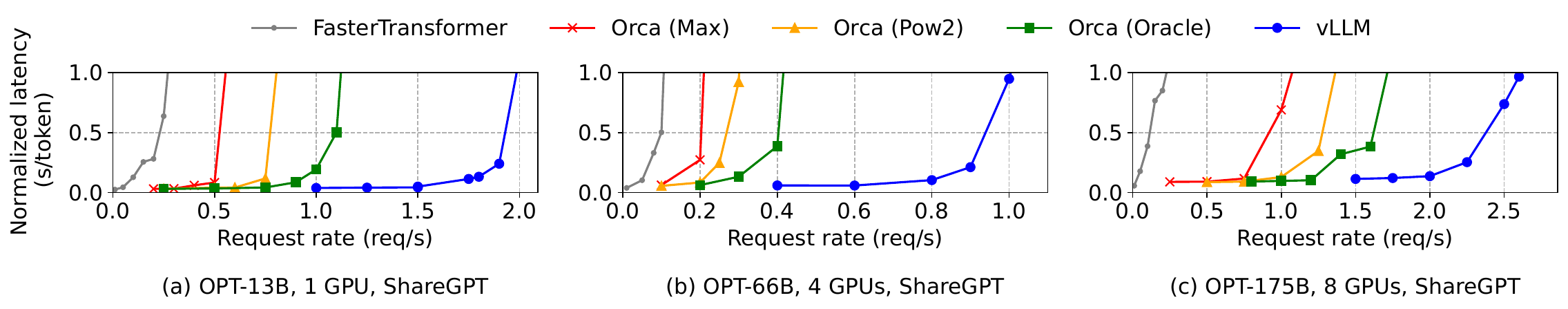}
    \includegraphics[scale=0.46]{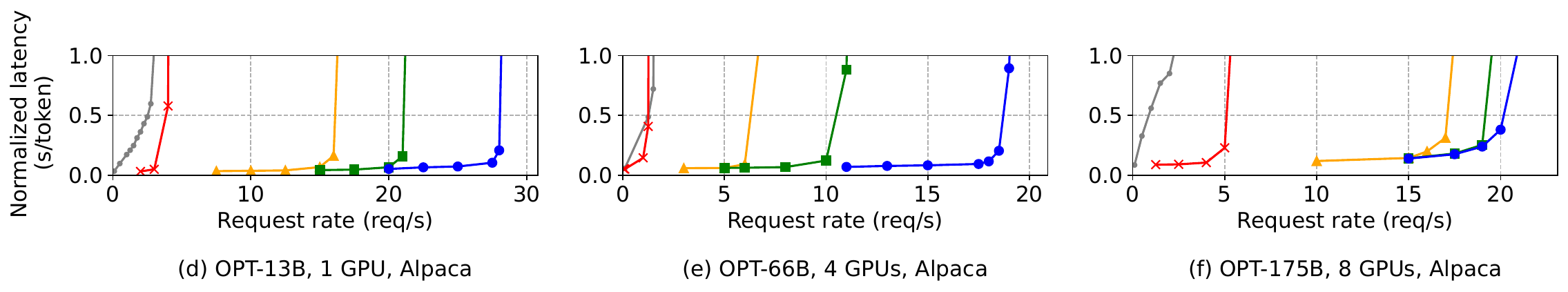}
\vspace{-8pt}
\caption{Single sequence generation with OPT models on the ShareGPT and Alpaca dataset}
\vspace{-10pt}
\label{fig:single-sequence}
\end{figure*}

\begin{figure}[t]
     \centering
     \begin{subfigure}[t]{0.48\linewidth}
         \centering
         \includegraphics[width=\columnwidth]{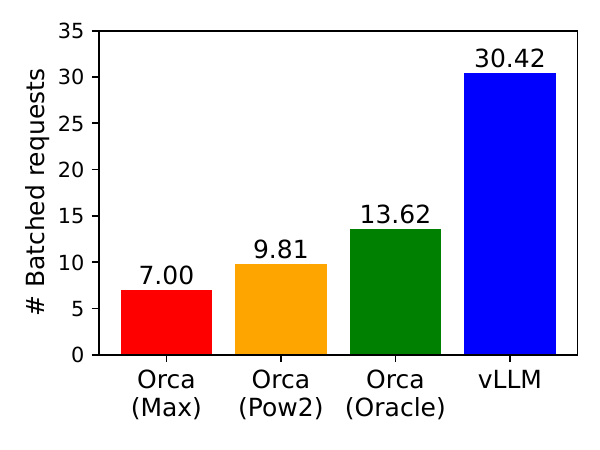}
         \vspace{-20pt}
         \caption{\small ShareGPT}
     \label{fig:batch-sharegpt}
     \end{subfigure}
     \begin{subfigure}[t]{0.48\linewidth}
         \centering
         \includegraphics[width=\columnwidth]{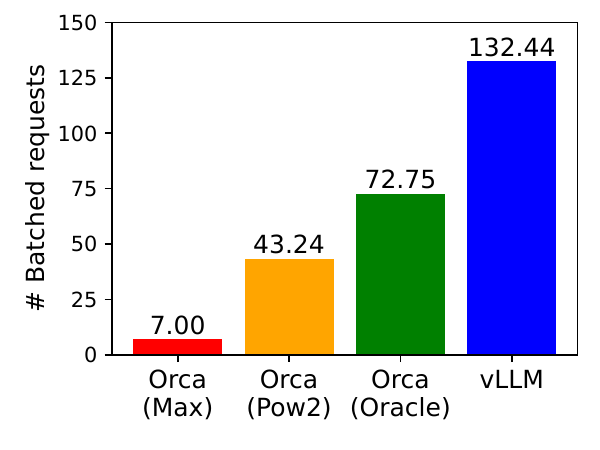}
         \vspace{-20pt}
         \caption{\small Alpaca}
     \label{fig:batch-alpaca}
     \end{subfigure}
     \vspace{-8pt}
     \caption{Average number of batched requests when serving OPT-13B for the ShareGPT (2 reqs/s) and Alpaca (30 reqs/s) traces.}
\vspace{-15pt}
\end{figure}

\section{Evaluation}
\label{sec:eval}

In this section, we evaluate the performance of \sys under a variety of workloads.

\subsection{Experimental Setup}
\label{subsec:exp-setup}

\heading{Model and server configurations.}
We use OPT~\cite{zhang2022opt} models with 13B, 66B, and 175B parameters and LLaMA~\cite{touvron2023llama} with 13B parameters for our evaluation.
13B and 66B are popular sizes for LLMs as shown in an LLM leaderboard~\cite{lmsysweek8}, while 175B is the size of the famous GPT-3~\cite{brown2020language} model.
For all of our experiments, we use A2 instances with NVIDIA A100 GPUs on Google Cloud Platform.
The detailed model sizes and server configurations are shown in Table~\ref{table:model_config}.

\heading{Workloads.}
We synthesize workloads based on ShareGPT~\cite{sharegpt} and Alpaca~\cite{alpaca} datasets, which contain input and output texts of real LLM services.
The ShareGPT dataset is a collection of user-shared conversations with ChatGPT~\cite{chatgpt}.
The Alpaca dataset is an instruction dataset generated by GPT-3.5 with self-instruct~\cite{wang2022self}.
We tokenize the datasets and use their input and output lengths to synthesize client requests.
As shown in Fig.~\ref{fig:dataset-length-dist}, the ShareGPT dataset has 8.4$\times$ longer input prompts and 5.8$\times$ longer outputs on average than the Alpaca dataset, with higher variance.
Since these datasets do not include timestamps, we generate request arrival times using Poisson distribution with different request rates.

\heading{Baseline 1: FasterTransformer.}
FasterTransformer~\cite{nvidiaft} is a distributed inference engine highly optimized for latency.
As FasterTransformer does not have its own scheduler, we implement a custom scheduler with a dynamic batching mechanism similar to the existing serving systems such as Triton~\cite{nvidiatriton}.
Specifically, we set a maximum batch size $B$ as large as possible for each experiment, according to the GPU memory capacity.
The scheduler takes up to $B$ number of earliest arrived requests and sends the batch to FasterTransformer for processing.

\begin{figure*}[t]
    \centering
    \includegraphics[scale=0.46]{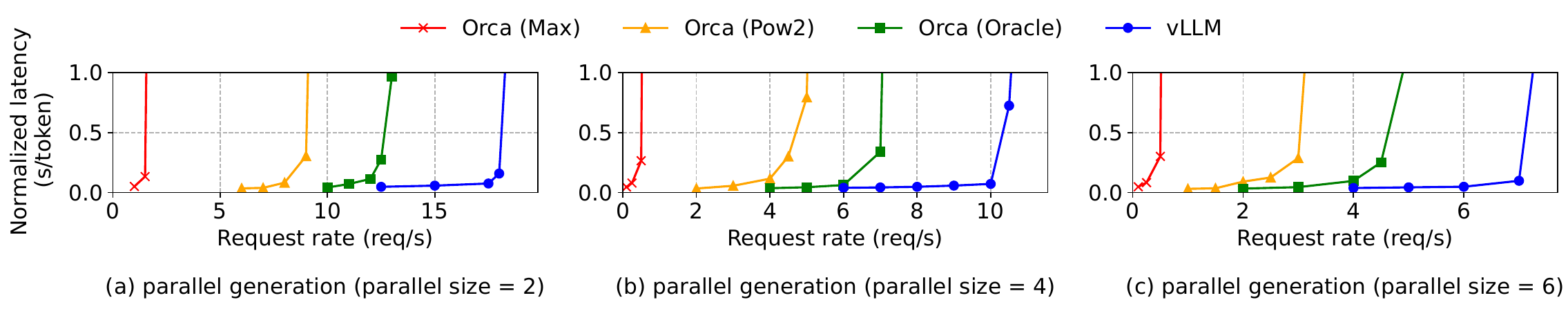}
    \includegraphics[scale=0.46]{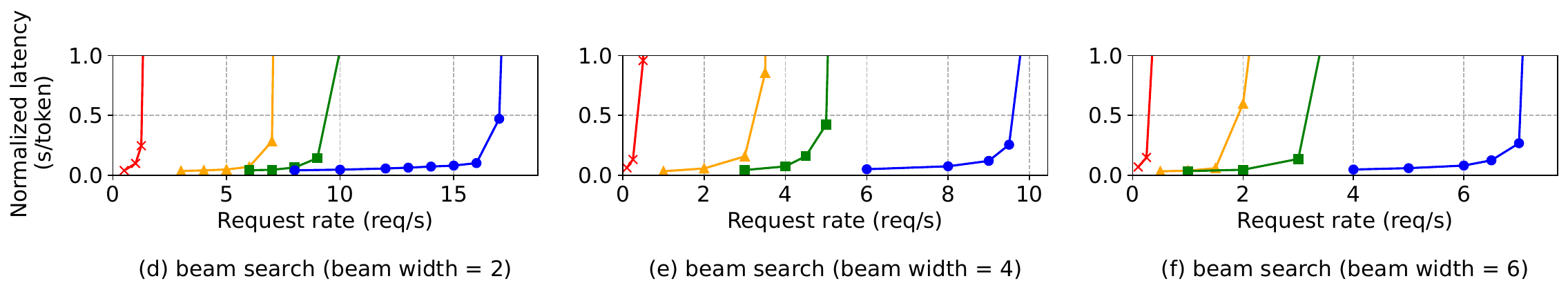}
\caption{Parallel generation and beam search with OPT-13B on the Alpaca dataset.}
\label{fig:parallel-beam-alpaca}
\vspace{-5pt}
\end{figure*}

\heading{Baseline 2: Orca.}
Orca~\cite{yu2022orca} is a state-of-the-art LLM serving system optimized for throughput.
Since Orca is not publicly available for use, we implement our own version of Orca.
We assume Orca uses the buddy allocation algorithm to determine the memory address to store KV cache.
We implement three versions of Orca based on how much it over-reserves the space for request outputs:
\begin{CompactItemize}
    \item \textbf{Orca (Oracle).} We assume the system has the knowledge of the lengths of the outputs that will be actually generated for the requests. This shows the upper-bound performance of Orca, which is infeasible to achieve in practice.
    \item \textbf{Orca (Pow2).} We assume the system over-reserves the space for outputs by at most 2$\times$. For example, if the true output length is 25, it reserves 32 positions for outputs.
    \item \textbf{Orca (Max).} We assume the system always reserves the space up to the maximum sequence length of the model, i.e., 2048 tokens.
\end{CompactItemize}

\heading{Key metrics.}
We focus on serving throughput.
Specifically, using the workloads with different request rates, we measure \textit{normalized latency} of the systems, the mean of every request's end-to-end latency divided by its output length, as in Orca \cite{yu2022orca}.
A high-throughput serving system should retain low normalized latency against high request rates.
For most experiments, we evaluate the systems with 1-hour traces.
As an exception, we use 15-minute traces for the OPT-175B model due to the cost limit.

\subsection{Basic Sampling}
\label{sec:eval:basic-sampling}

We evaluate the performance of \sys with basic sampling (one sample per request) on three models and two datasets.
The first row of Fig.~\ref{fig:single-sequence} shows the results on the ShareGPT dataset.
The curves illustrate that as the request rate increases, the latency initially increases at a gradual pace but then suddenly explodes.
This can be attributed to the fact that when the request rate surpasses the capacity of the serving system, the queue length continues to grow infinitely and so does the latency of the requests.

On the ShareGPT dataset, \sys can sustain $1.7\times$--$2.7\times$ higher request rates compared to Orca (Oracle) and $2.7\times$--$8\times$ compared to Orca (Max), while maintaining similar latencies.
This is because \sys's \tech can efficiently manage the memory usage and thus enable batching more requests than Orca.
For example, as shown in Fig.~\ref{fig:batch-sharegpt}, for OPT-13B vLLM processes $2.2\times$ more requests at the same time than Orca (Oracle) and $4.3\times$ more requests than Orca (Max).
Compared to FasterTransformer, \sys can sustain up to $22\times$ higher request rates, as FasterTransformer does not utilize a fine-grained scheduling mechanism and inefficiently manages the memory like Orca (Max).

The second row of Fig.~\ref{fig:single-sequence} and Fig.~\ref{fig:batch-alpaca} shows the results on the Alpaca dataset, which follows a similar trend to the ShareGPT dataset.
One exception is Fig.~\ref{fig:single-sequence} (f), where \sys's advantage over Orca (Oracle) and Orca (Pow2) is less pronounced.
This is because the model and server configuration for OPT-175B (Table~\ref{table:model_config}) allows for large GPU memory space available to store KV cache, while the Alpaca dataset has short sequences.
In this setup, Orca (Oracle) and Orca (Pow2) can also batch a large number of requests despite the inefficiencies in their memory management.
As a result, the performance of the systems becomes compute-bound rather than memory-bound.

\begin{figure}[t]
     \centering
     \begin{subfigure}[b]{0.5\linewidth}
         \centering
         \includegraphics[width=.9\columnwidth]{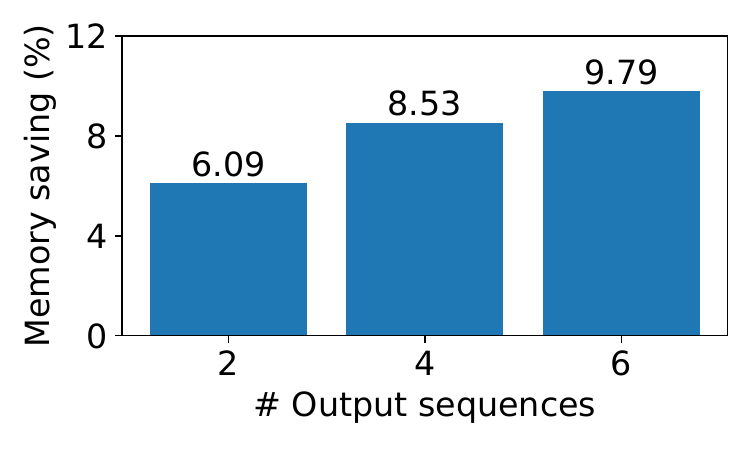}
         \vspace{-5pt}
         \caption{Parallel sampling}
     \end{subfigure}
     \begin{subfigure}[b]{0.5\linewidth}
         \centering
         \includegraphics[width=.9\columnwidth]{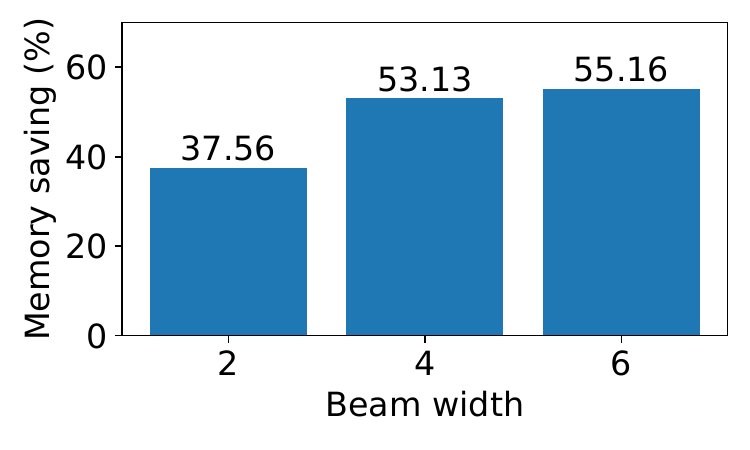}
         \vspace{-5pt}
         \caption{Beam search}
     \end{subfigure}
     \vspace{-7pt}
     \caption{Average amount of memory saving from sharing KV blocks, when serving OPT-13B for the Alpaca trace.}
    \vspace{-12pt}
\label{fig:memory-saving-sharing}
\end{figure}

\subsection{Parallel Sampling and Beam Search}
\label{sec:eval:beamsearch}

We evaluate the effectiveness of memory sharing in \tech with two popular sampling methods: parallel sampling and beam search.
In parallel sampling, all parallel sequences in a request can share the KV cache for the prompt.
As shown in the first row of Fig.~\ref{fig:parallel-beam-alpaca}, with a larger number of sequences to sample, \sys brings more improvement over the Orca baselines.
Similarly, the second row of Fig.~\ref{fig:parallel-beam-alpaca} shows the results for beam search with different beam widths.
Since beam search allows for more sharing, \sys demonstrates even greater performance benefits.
The improvement of \sys over Orca (Oracle) on OPT-13B and the Alpaca dataset goes from $1.3 \times$ in basic sampling to $2.3 \times$ in beam search with a width of 6.

Fig.~\ref{fig:memory-saving-sharing} plots the amount of memory saving, computed by the number of blocks we saved by sharing divided by the number of total blocks without sharing. We show 6.1\% - 9.8\% memory saving on parallel sampling and 37.6\% - 55.2\% on beam search.
In the same experiments with the ShareGPT dataset, we saw 16.2\% - 30.5\% memory saving on parallel sampling and 44.3\% - 66.3\% on beam search.

\subsection{Shared prefix}

\begin{figure}[t]
    \centering
    \includegraphics[width=.9\columnwidth]{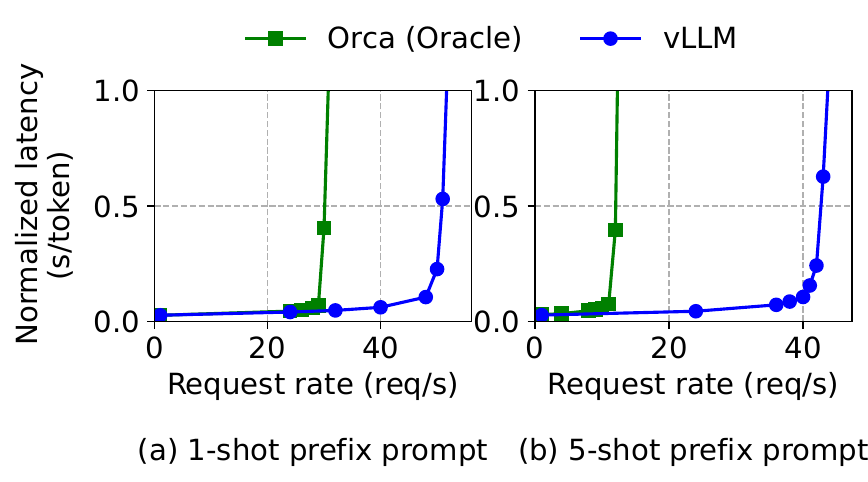}
    \vspace{-5pt}
    \caption{Translation workload where the input prompts share a common prefix. The prefix includes (a) 1 example with 80 tokens or (b) 5 examples with 341 tokens.}
\label{fig:prefix-share-exp}
\end{figure}

We explore the effectiveness of \sys for the case a prefix is shared among different input prompts, as illustrated in Fig.~\ref{fig:share-prompt}.
For the model, we use LLaMA-13B~\cite{touvron2023llama}, which is multilingual.
For the workload, we use the WMT16~\cite{bojar-EtAl:2016:WMT1} English-to-German translation dataset and synthesize two prefixes that include an instruction and a few translation examples.
The first prefix includes a single example (i.e., one-shot) while the other prefix includes 5 examples (i.e., few-shot).
As shown in Fig.~\ref{fig:prefix-share-exp} (a), \sys achieves $1.67 \times$ higher throughput than Orca (Oracle) when the one-shot prefix is shared.
Furthermore, when more examples are shared (Fig.~\ref{fig:prefix-share-exp} (b)), \sys achieves $3.58 \times$ higher throughput than Orca (Oracle).

\begin{figure}
    \centering
    \includegraphics[width=0.8\columnwidth]{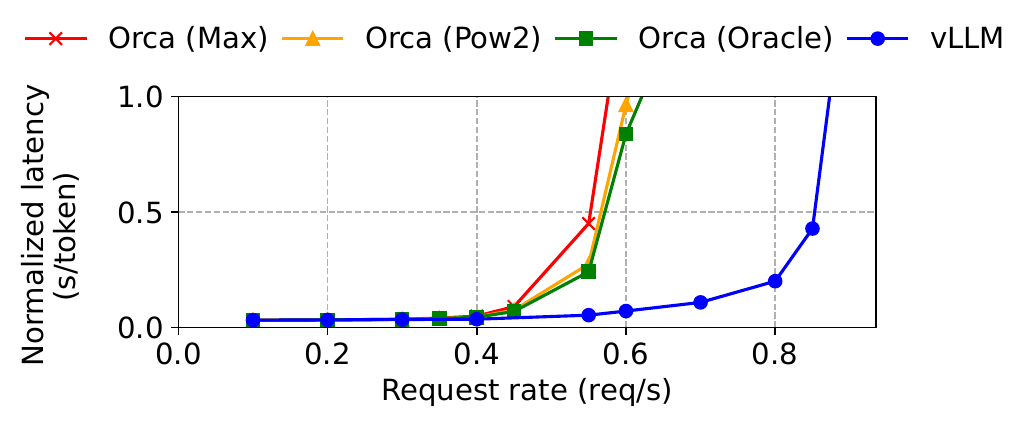}
    \vspace{-7pt}
    \caption{Performance on chatbot workload.}
    \label{fig:chatbot}
\end{figure}

\subsection{Chatbot}
\label{sec:chatbot}
A chatbot \cite{chatgpt, bard, vicuna2023} is one of the most important applications of LLMs.
To implement a chatbot, we let the model generate a response by concatenating the chatting history and the last user query into a prompt.
We synthesize the chatting history and user query using the ShareGPT dataset.
Due to the limited context length of the OPT-13B model, we cut the prompt to the last 1024 tokens and let the model generate at most 1024 tokens.
We do not store the KV cache between different conversation rounds as doing this would occupy the space for other requests between the conversation rounds.

Fig.~\ref{fig:chatbot} shows that \sys can sustain $2\times$ higher request rates compared to the three Orca baselines.
Since the ShareGPT dataset contains many long conversations, the input prompts for most requests have 1024 tokens.
Due to the buddy allocation algorithm, the Orca baselines reserve the space for 1024 tokens for the request outputs, regardless of how they predict the output lengths.
For this reason, the three Orca baselines behave similarly.
In contrast, \sys can effectively handle the long prompts, as \tech resolves the problem of memory fragmentation and reservation.

\section{Ablation Studies}

In this section, we study various aspects of \sys and evaluate the design choices we make with ablation experiments.

\subsection{Kernel Microbenchmark}

\begin{figure}[t]
     \centering
     \begin{subfigure}[t]{0.48\linewidth}
         \centering
         \includegraphics[width=.95\columnwidth]{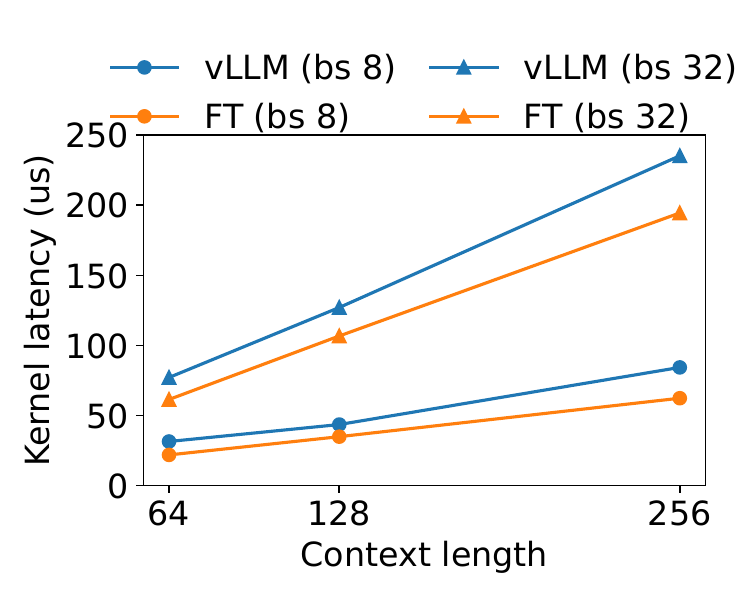}
         \caption{\small Latency of attention kernels.\label{fig:kernel-latency}}
     \end{subfigure}\hfill
     \begin{subfigure}[t]{0.48\linewidth}
         \centering
         \includegraphics[width=.9\columnwidth]{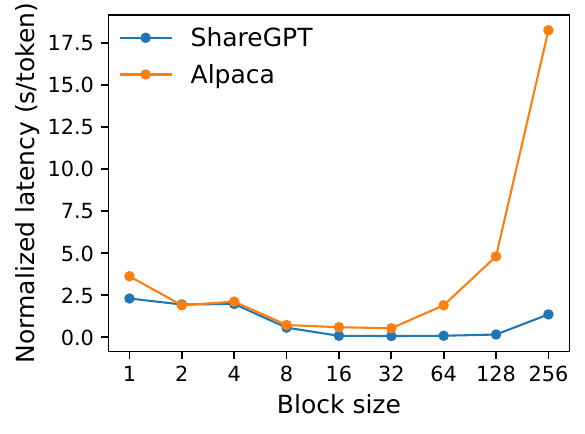}
         \caption{\small End-to-end latency with different block sizes. \label{fig:block-size-n1}}
     \end{subfigure}
     \vspace{-10pt}
     \caption{Ablation experiments.}
\end{figure}

The dynamic block mapping in \tech affects the performance of the GPU operations involving the stored KV cache, i.e., block read/writes and attention.
Compared to the existing systems, our GPU kernels (\S\ref{sec:impl}) involve extra overheads of accessing the block table, executing extra branches, and handling variable sequence lengths.
As shown in Fig.~\ref{fig:kernel-latency}, this leads to 20--26\% higher attention kernel latency, compared to the highly-optimized FasterTransformer implementation.
We believe the overhead is small as it only affects the attention operator but not the other operators in the model, such as Linear.
Despite the overhead, \tech makes \sys significantly outperform FasterTransformer in end-to-end performance (\S\ref{sec:eval}).

\subsection{Impact of Block Size}
\label{sec:eval:blocksize}

The choice of block size can have a substantial impact on the performance of \sys.
If the block size is too small, \sys may not fully utilize the GPU's parallelism for reading and processing KV cache.
If the block size is too large, internal fragmentation increases and the probability of sharing decreases.

In Fig.~\ref{fig:block-size-n1}, we evaluate the performance of \sys with different block sizes, using the ShareGPT and Alpaca traces with basic sampling under fixed request rates.
In the ShareGPT trace, block sizes from 16 to 128 lead to the best performance.
In the Alpaca trace, while the block size 16 and 32 work well, larger block sizes significantly degrade the performance since the sequences become shorter than the block sizes.
In practice, we find that the block size 16 is large enough to efficiently utilize the GPU and small enough to avoid significant internal fragmentation in most workloads.
Accordingly, \sys sets its default block size as 16.

\begin{figure}[t]
     \centering
     \begin{subfigure}[t]{0.48\linewidth}
         \centering         \includegraphics[width=.9\columnwidth]{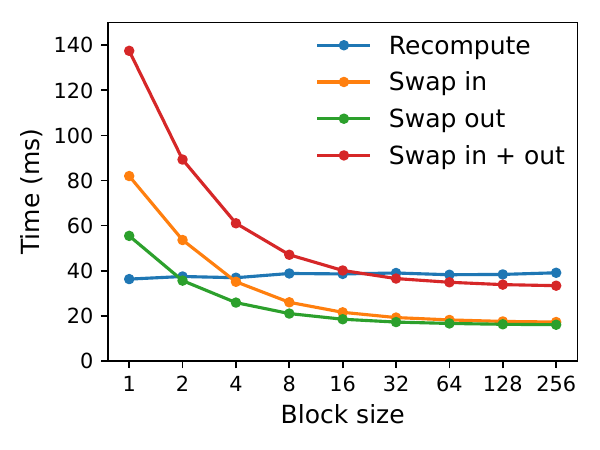}
         \vspace{-7pt}
         \caption{Microbenchmark\label{fig:recomp-vs-swap-micro}}
         
     \end{subfigure}
     \begin{subfigure}[t]{0.48\linewidth}
         \centering        \includegraphics[width=.9\columnwidth]{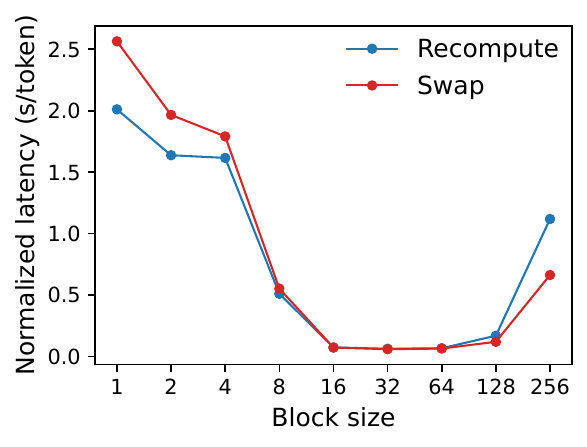}
         \vspace{-7pt}
         \caption{End-to-end performance\label{fig:recomp-vs-swap-e2e}}
     \end{subfigure}
     \vspace{-3pt}
     \caption{(a) Overhead of recomputation and swapping for different block sizes. (b) Performance when serving OPT-13B with the ShareGPT traces at the same request rate.}
     \vspace{-5pt}
\label{fig:recompute-vs-swap}
\end{figure}

\subsection{Comparing Recomputation and Swapping}
\label{sec:eval:scheduling}

\sys supports both recomputation and swapping as its recovery mechanisms.
To understand the tradeoffs between the two methods, we evaluate their end-to-end performance and microbenchmark their overheads, as presented in Fig.~\ref{fig:recompute-vs-swap}.
Our results reveal that swapping incurs excessive overhead with small block sizes.
This is because small block sizes often result in numerous small data transfers between CPU and GPU, which limits the effective PCIe bandwidth.
In contrast, the overhead of recomputation remains constant across different block sizes, as recomputation does not utilize the KV blocks.
Thus, recomputation is more efficient when the block size is small, while swapping is more efficient when the block size is large, though recomputation overhead is never higher than 20\% of swapping's latency.
For medium block sizes from 16 to 64, the two methods exhibit comparable end-to-end performance.

\section{Discussion}

\heading{Applying the virtual memory and paging technique to other GPU workloads.}
The idea of virtual memory and paging is effective for managing the KV cache in LLM serving because the workload requires dynamic memory allocation (since the output length is not known a priori) and its performance is bound by the GPU memory capacity.
However, this does not generally hold for every GPU workload.
For example, in DNN training, the tensor shapes are typically static, and thus memory allocation can be optimized ahead of time.
For another example, in serving DNNs that are not LLMs, an increase in memory efficiency may not result in any performance improvement since the performance is primarily compute-bound.
In such scenarios, introducing the \sys's techniques may rather degrade the performance due to the extra overhead of memory indirection and non-contiguous block memory.
However, we would be excited to see \sys's techniques being applied to other workloads with similar properties to LLM serving.

\heading{LLM-specific optimizations in applying virtual memory and paging.}
\sys re-interprets and augments the idea of virtual memory and paging by leveraging the application-specific semantics.
One example is \sys's all-or-nothing swap-out policy, which exploits the fact that processing a request requires all of its corresponding token states to be stored in GPU memory.
Another example is the recomputation method to recover the evicted blocks, which is not feasible in OS.
Besides, \sys mitigates the overhead of memory indirection in paging by fusing the GPU kernels for memory access operations with those for other operations such as attention.

\section{Related Work}
\label{sec:related_work}

\heading{General model serving systems.}
Model serving has been an active area of research in recent years, with numerous systems proposed to tackle diverse aspects of deep learning model deployment.
Clipper~\cite{crankshaw2017clipper}, TensorFlow Serving~\cite{olston2017tensorflow}, Nexus~\cite{shen2019nexus}, InferLine~\cite{crankshaw2020inferline}, and
Clockwork~\cite{gujarati2020serving} are some earlier general model serving systems. They study batching, caching, placement, and scheduling for serving single or multiple models.
More recently, DVABatch~\cite{cui2022dvabatch} introduces multi-entry multi-exit batching. REEF~\cite{han2022microsecond} and  Shepherd~\cite{zhang2023shepherd} propose preemption for serving. AlpaServe~\cite{li2023alpaserve} utilizes model parallelism for statistical multiplexing.
However, these general systems fail to take into account the auto-regressive property and token state of LLM inference, resulting in missed opportunities for optimization.

\heading{Specialized serving systems for transformers.}
Due to the significance of the transformer architecture, numerous specialized serving systems for it have been developed. These systems utilize GPU kernel optimizations~\cite{wang2021lightseq,aminabadi2022deepspeed,nvidiaft,ma2020rammer},
advanced batching mechanisms~\cite{fang2021turbotransformers,yu2022orca},
model parallelism~\cite{pope2022efficiently,yu2022orca,aminabadi2022deepspeed}, and parameter sharing~\cite{zhou2022pets} for efficient serving.
Among them, Orca~\cite{yu2022orca} is most relevant to our approach.

\heading{Comparison to Orca.}
The iteration-level scheduling in Orca~\cite{yu2022orca} and \tech in \sys are complementary techniques: While both systems aim to increase the GPU utilization and hence the throughput of LLM serving, Orca achieves it by scheduling and interleaving the requests so that more requests can be processed in parallel, while \sys is doing so by increasing memory utilization so that the working sets of more requests fit into memory.
By reducing memory fragmentation and enabling sharing, \sys runs more requests in a batch in parallel and achieves a 2-4$\times$ speedup compared to Orca.
Indeed, the fine-grained scheduling and interleaving of the requests like in Orca makes memory management more challenging, making the techniques proposed in \sys even more crucial.

\heading{Memory optimizations.}
The widening gap between the compute capability and memory capacity of accelerators has caused memory to become a bottleneck for both training and inference.
Swapping~\cite{huang2020swapadvisor,wang2018superneurons,ren2021zero}, recomputation~\cite{chen2016training,jain2020checkmate} and their combination~\cite{patil2022poet} have been utilized to reduce the peak memory of training.
Notably, FlexGen~\cite{sheng2023high} studies how to swap weights and token states for LLM inference with limited GPU memory, but it does not target the online serving settings.
OLLA~\cite{steiner2022olla} optimizes the lifetime and location of tensors to reduce fragmentation, but it does not do fine-grained block-level management or online serving.
FlashAttention~\cite{dao2022flashattention} applies tiling and kernel optimizations to reduce the peak memory of attention computation and reduce I/O costs.
This paper introduces a new idea of block-level memory management in the context of online serving.

\section{Conclusion}
This paper proposes \tech, a new attention algorithm that allows attention keys and values to be stored in non-contiguous paged memory, and presents \sys, a high-throughput LLM serving system with efficient memory management enabled by \tech.
Inspired by operating systems, we demonstrate how established techniques, such as virtual memory and copy-on-write, can be adapted to efficiently manage KV cache and handle various decoding algorithms in LLM serving.
Our experiments show that \sys achieves 2-4$\times$ throughput improvements over the state-of-the-art systems.

\section*{Acknowledgement}
We would like to thank Xiaoxuan Liu, Zhifeng Chen, Yanping Huang, anonymous SOSP reviewers, and our shepherd, Lidong Zhou, for their insightful feedback.
This research is partly supported by gifts from Andreessen Horowitz, Anyscale, Astronomer, Google, IBM, Intel, Lacework, Microsoft, Mohamed Bin Zayed University of Artificial Intelligence, Samsung SDS, Uber, and VMware.

\bibliographystyle{ACM-Reference-Format}
\bibliography{reference}


\begin{thebibliography}{64}


\ifx \showCODEN    \undefined \def \showCODEN     #1{\unskip}     \fi
\ifx \showDOI      \undefined \def \showDOI       #1{#1}\fi
\ifx \showISBNx    \undefined \def \showISBNx     #1{\unskip}     \fi
\ifx \showISBNxiii \undefined \def \showISBNxiii  #1{\unskip}     \fi
\ifx \showISSN     \undefined \def \showISSN      #1{\unskip}     \fi
\ifx \showLCCN     \undefined \def \showLCCN      #1{\unskip}     \fi
\ifx \shownote     \undefined \def \shownote      #1{#1}          \fi
\ifx \showarticletitle \undefined \def \showarticletitle #1{#1}   \fi
\ifx \showURL      \undefined \def \showURL       {\relax}        \fi
\providecommand\bibfield[2]{#2}
\providecommand\bibinfo[2]{#2}
\providecommand\natexlab[1]{#1}
\providecommand\showeprint[2][]{arXiv:#2}

\bibitem[Aminabadi et~al\mbox{.}(2022)]%
        {aminabadi2022deepspeed}
\bibfield{author}{\bibinfo{person}{Reza~Yazdani Aminabadi}, \bibinfo{person}{Samyam Rajbhandari}, \bibinfo{person}{Minjia Zhang}, \bibinfo{person}{Ammar~Ahmad Awan}, \bibinfo{person}{Cheng Li}, \bibinfo{person}{Du Li}, \bibinfo{person}{Elton Zheng}, \bibinfo{person}{Jeff Rasley}, \bibinfo{person}{Shaden Smith}, \bibinfo{person}{Olatunji Ruwase}, {et~al\mbox{.}}} \bibinfo{year}{2022}\natexlab{}.
\newblock \showarticletitle{DeepSpeed Inference: Enabling Efficient Inference of Transformer Models at Unprecedented Scale}.
\newblock \bibinfo{journal}{\emph{arXiv preprint arXiv:2207.00032}} (\bibinfo{year}{2022}).
\newblock


\bibitem[Ba et~al\mbox{.}(2016)]%
        {ba2016layer}
\bibfield{author}{\bibinfo{person}{Jimmy~Lei Ba}, \bibinfo{person}{Jamie~Ryan Kiros}, {and} \bibinfo{person}{Geoffrey~E Hinton}.} \bibinfo{year}{2016}\natexlab{}.
\newblock \showarticletitle{Layer normalization}.
\newblock \bibinfo{journal}{\emph{arXiv preprint arXiv:1607.06450}} (\bibinfo{year}{2016}).
\newblock


\bibitem[Bengio et~al\mbox{.}(2000)]%
        {bengio2000neural}
\bibfield{author}{\bibinfo{person}{Yoshua Bengio}, \bibinfo{person}{R{\'e}jean Ducharme}, {and} \bibinfo{person}{Pascal Vincent}.} \bibinfo{year}{2000}\natexlab{}.
\newblock \showarticletitle{A neural probabilistic language model}.
\newblock \bibinfo{journal}{\emph{Advances in neural information processing systems}}  \bibinfo{volume}{13} (\bibinfo{year}{2000}).
\newblock


\bibitem[Bojar et~al\mbox{.}(2016)]%
        {bojar-EtAl:2016:WMT1}
\bibfield{author}{\bibinfo{person}{Ond~{r}ej Bojar}, \bibinfo{person}{Rajen Chatterjee}, \bibinfo{person}{Christian Federmann}, \bibinfo{person}{Yvette Graham}, \bibinfo{person}{Barry Haddow}, \bibinfo{person}{Matthias Huck}, \bibinfo{person}{Antonio Jimeno~Yepes}, \bibinfo{person}{Philipp Koehn}, \bibinfo{person}{Varvara Logacheva}, \bibinfo{person}{Christof Monz}, \bibinfo{person}{Matteo Negri}, \bibinfo{person}{Aurelie Neveol}, \bibinfo{person}{Mariana Neves}, \bibinfo{person}{Martin Popel}, \bibinfo{person}{Matt Post}, \bibinfo{person}{Raphael Rubino}, \bibinfo{person}{Carolina Scarton}, \bibinfo{person}{Lucia Specia}, \bibinfo{person}{Marco Turchi}, \bibinfo{person}{Karin Verspoor}, {and} \bibinfo{person}{Marcos Zampieri}.} \bibinfo{year}{2016}\natexlab{}.
\newblock \showarticletitle{Findings of the 2016 Conference on Machine Translation}. In \bibinfo{booktitle}{\emph{Proceedings of the First Conference on Machine Translation}}. \bibinfo{publisher}{Association for Computational Linguistics}, \bibinfo{address}{Berlin, Germany}, \bibinfo{pages}{131--198}.
\newblock
\urldef\tempurl%
\url{http://www.aclweb.org/anthology/W/W16/W16-2301}
\showURL{%
\tempurl}


\bibitem[Brown et~al\mbox{.}(2020)]%
        {brown2020language}
\bibfield{author}{\bibinfo{person}{Tom Brown}, \bibinfo{person}{Benjamin Mann}, \bibinfo{person}{Nick Ryder}, \bibinfo{person}{Melanie Subbiah}, \bibinfo{person}{Jared~D Kaplan}, \bibinfo{person}{Prafulla Dhariwal}, \bibinfo{person}{Arvind Neelakantan}, \bibinfo{person}{Pranav Shyam}, \bibinfo{person}{Girish Sastry}, \bibinfo{person}{Amanda Askell}, {et~al\mbox{.}}} \bibinfo{year}{2020}\natexlab{}.
\newblock \showarticletitle{Language models are few-shot learners}.
\newblock \bibinfo{journal}{\emph{Advances in neural information processing systems}}  \bibinfo{volume}{33} (\bibinfo{year}{2020}), \bibinfo{pages}{1877--1901}.
\newblock


\bibitem[Chen et~al\mbox{.}(2021)]%
        {chen2021evaluating}
\bibfield{author}{\bibinfo{person}{Mark Chen}, \bibinfo{person}{Jerry Tworek}, \bibinfo{person}{Heewoo Jun}, \bibinfo{person}{Qiming Yuan}, \bibinfo{person}{Henrique Ponde de~Oliveira Pinto}, \bibinfo{person}{Jared Kaplan}, \bibinfo{person}{Harri Edwards}, \bibinfo{person}{Yuri Burda}, \bibinfo{person}{Nicholas Joseph}, \bibinfo{person}{Greg Brockman}, {et~al\mbox{.}}} \bibinfo{year}{2021}\natexlab{}.
\newblock \showarticletitle{Evaluating large language models trained on code}.
\newblock \bibinfo{journal}{\emph{arXiv preprint arXiv:2107.03374}} (\bibinfo{year}{2021}).
\newblock


\bibitem[Chen et~al\mbox{.}(2016)]%
        {chen2016training}
\bibfield{author}{\bibinfo{person}{Tianqi Chen}, \bibinfo{person}{Bing Xu}, \bibinfo{person}{Chiyuan Zhang}, {and} \bibinfo{person}{Carlos Guestrin}.} \bibinfo{year}{2016}\natexlab{}.
\newblock \showarticletitle{Training deep nets with sublinear memory cost}.
\newblock \bibinfo{journal}{\emph{arXiv preprint arXiv:1604.06174}} (\bibinfo{year}{2016}).
\newblock


\bibitem[Chiang et~al\mbox{.}(2023)]%
        {vicuna2023}
\bibfield{author}{\bibinfo{person}{Wei-Lin Chiang}, \bibinfo{person}{Zhuohan Li}, \bibinfo{person}{Zi Lin}, \bibinfo{person}{Ying Sheng}, \bibinfo{person}{Zhanghao Wu}, \bibinfo{person}{Hao Zhang}, \bibinfo{person}{Lianmin Zheng}, \bibinfo{person}{Siyuan Zhuang}, \bibinfo{person}{Yonghao Zhuang}, \bibinfo{person}{Joseph~E. Gonzalez}, \bibinfo{person}{Ion Stoica}, {and} \bibinfo{person}{Eric~P. Xing}.} \bibinfo{year}{2023}\natexlab{}.
\newblock \bibinfo{title}{Vicuna: An Open-Source Chatbot Impressing GPT-4 with 90\%* ChatGPT Quality}.
\newblock
\newblock
\urldef\tempurl%
\url{https://lmsys.org/blog/2023-03-30-vicuna/}
\showURL{%
\tempurl}


\bibitem[Chowdhery et~al\mbox{.}(2022)]%
        {chowdhery2022palm}
\bibfield{author}{\bibinfo{person}{Aakanksha Chowdhery}, \bibinfo{person}{Sharan Narang}, \bibinfo{person}{Jacob Devlin}, \bibinfo{person}{Maarten Bosma}, \bibinfo{person}{Gaurav Mishra}, \bibinfo{person}{Adam Roberts}, \bibinfo{person}{Paul Barham}, \bibinfo{person}{Hyung~Won Chung}, \bibinfo{person}{Charles Sutton}, \bibinfo{person}{Sebastian Gehrmann}, {et~al\mbox{.}}} \bibinfo{year}{2022}\natexlab{}.
\newblock \showarticletitle{Palm: Scaling language modeling with pathways}.
\newblock \bibinfo{journal}{\emph{arXiv preprint arXiv:2204.02311}} (\bibinfo{year}{2022}).
\newblock


\bibitem[Crankshaw et~al\mbox{.}(2020)]%
        {crankshaw2020inferline}
\bibfield{author}{\bibinfo{person}{Daniel Crankshaw}, \bibinfo{person}{Gur-Eyal Sela}, \bibinfo{person}{Xiangxi Mo}, \bibinfo{person}{Corey Zumar}, \bibinfo{person}{Ion Stoica}, \bibinfo{person}{Joseph Gonzalez}, {and} \bibinfo{person}{Alexey Tumanov}.} \bibinfo{year}{2020}\natexlab{}.
\newblock \showarticletitle{InferLine: latency-aware provisioning and scaling for prediction serving pipelines}. In \bibinfo{booktitle}{\emph{Proceedings of the 11th ACM Symposium on Cloud Computing}}. \bibinfo{pages}{477--491}.
\newblock


\bibitem[Crankshaw et~al\mbox{.}(2017)]%
        {crankshaw2017clipper}
\bibfield{author}{\bibinfo{person}{Daniel Crankshaw}, \bibinfo{person}{Xin Wang}, \bibinfo{person}{Guilio Zhou}, \bibinfo{person}{Michael~J Franklin}, \bibinfo{person}{Joseph~E Gonzalez}, {and} \bibinfo{person}{Ion Stoica}.} \bibinfo{year}{2017}\natexlab{}.
\newblock \showarticletitle{Clipper: A Low-Latency Online Prediction Serving System}. In \bibinfo{booktitle}{\emph{14th USENIX Symposium on Networked Systems Design and Implementation (NSDI 17)}}. \bibinfo{pages}{613--627}.
\newblock


\bibitem[Cui et~al\mbox{.}(2022)]%
        {cui2022dvabatch}
\bibfield{author}{\bibinfo{person}{Weihao Cui}, \bibinfo{person}{Han Zhao}, \bibinfo{person}{Quan Chen}, \bibinfo{person}{Hao Wei}, \bibinfo{person}{Zirui Li}, \bibinfo{person}{Deze Zeng}, \bibinfo{person}{Chao Li}, {and} \bibinfo{person}{Minyi Guo}.} \bibinfo{year}{2022}\natexlab{}.
\newblock \showarticletitle{DVABatch: Diversity-aware Multi-Entry Multi-Exit Batching for Efficient Processing of DNN Services on GPUs}. In \bibinfo{booktitle}{\emph{2022 USENIX Annual Technical Conference (USENIX ATC 22)}}. \bibinfo{pages}{183--198}.
\newblock


\bibitem[Dao et~al\mbox{.}(2022)]%
        {dao2022flashattention}
\bibfield{author}{\bibinfo{person}{Tri Dao}, \bibinfo{person}{Dan Fu}, \bibinfo{person}{Stefano Ermon}, \bibinfo{person}{Atri Rudra}, {and} \bibinfo{person}{Christopher R{\'e}}.} \bibinfo{year}{2022}\natexlab{}.
\newblock \showarticletitle{Flashattention: Fast and memory-efficient exact attention with io-awareness}.
\newblock \bibinfo{journal}{\emph{Advances in Neural Information Processing Systems}}  \bibinfo{volume}{35} (\bibinfo{year}{2022}), \bibinfo{pages}{16344--16359}.
\newblock


\bibitem[Fang et~al\mbox{.}(2021)]%
        {fang2021turbotransformers}
\bibfield{author}{\bibinfo{person}{Jiarui Fang}, \bibinfo{person}{Yang Yu}, \bibinfo{person}{Chengduo Zhao}, {and} \bibinfo{person}{Jie Zhou}.} \bibinfo{year}{2021}\natexlab{}.
\newblock \showarticletitle{TurboTransformers: an efficient GPU serving system for transformer models}. In \bibinfo{booktitle}{\emph{Proceedings of the 26th ACM SIGPLAN Symposium on Principles and Practice of Parallel Programming}}. \bibinfo{pages}{389--402}.
\newblock


\bibitem[FastAPI(2023)]%
        {fastapi}
\bibfield{author}{\bibinfo{person}{FastAPI}.} \bibinfo{year}{2023}\natexlab{}.
\newblock \bibinfo{title}{FastAPI}.
\newblock \bibinfo{howpublished}{\url{https://github.com/tiangolo/fastapi}}.
\newblock


\bibitem[Gao et~al\mbox{.}(2018)]%
        {gao2018low}
\bibfield{author}{\bibinfo{person}{Pin Gao}, \bibinfo{person}{Lingfan Yu}, \bibinfo{person}{Yongwei Wu}, {and} \bibinfo{person}{Jinyang Li}.} \bibinfo{year}{2018}\natexlab{}.
\newblock \showarticletitle{Low latency rnn inference with cellular batching}. In \bibinfo{booktitle}{\emph{Proceedings of the Thirteenth EuroSys Conference}}. \bibinfo{pages}{1--15}.
\newblock


\bibitem[Gholami et~al\mbox{.}(2021)]%
        {gholami2021ai}
\bibfield{author}{\bibinfo{person}{Amir Gholami}, \bibinfo{person}{Zhewei Yao}, \bibinfo{person}{Sehoon Kim}, \bibinfo{person}{Michael~W Mahoney}, {and} \bibinfo{person}{Kurt Keutzer}.} \bibinfo{year}{2021}\natexlab{}.
\newblock \showarticletitle{Ai and memory wall}.
\newblock \bibinfo{journal}{\emph{RiseLab Medium Post}}  \bibinfo{volume}{1} (\bibinfo{year}{2021}), \bibinfo{pages}{6}.
\newblock


\bibitem[Github(2022)]%
        {copilot}
\bibfield{author}{\bibinfo{person}{Github}.} \bibinfo{year}{2022}\natexlab{}.
\newblock
\newblock
\urldef\tempurl%
\url{https://github.com/features/copilot}
\showURL{%
\tempurl}


\bibitem[Google(2023)]%
        {bard}
\bibfield{author}{\bibinfo{person}{Google}.} \bibinfo{year}{2023}\natexlab{}.
\newblock
\newblock
\urldef\tempurl%
\url{https://bard.google.com/}
\showURL{%
\tempurl}


\bibitem[Gujarati et~al\mbox{.}(2020)]%
        {gujarati2020serving}
\bibfield{author}{\bibinfo{person}{Arpan Gujarati}, \bibinfo{person}{Reza Karimi}, \bibinfo{person}{Safya Alzayat}, \bibinfo{person}{Wei Hao}, \bibinfo{person}{Antoine Kaufmann}, \bibinfo{person}{Ymir Vigfusson}, {and} \bibinfo{person}{Jonathan Mace}.} \bibinfo{year}{2020}\natexlab{}.
\newblock \showarticletitle{Serving $\{$DNNs$\}$ like Clockwork: Performance Predictability from the Bottom Up}. In \bibinfo{booktitle}{\emph{14th USENIX Symposium on Operating Systems Design and Implementation (OSDI 20)}}. \bibinfo{pages}{443--462}.
\newblock


\bibitem[Han et~al\mbox{.}(2022)]%
        {han2022microsecond}
\bibfield{author}{\bibinfo{person}{Mingcong Han}, \bibinfo{person}{Hanze Zhang}, \bibinfo{person}{Rong Chen}, {and} \bibinfo{person}{Haibo Chen}.} \bibinfo{year}{2022}\natexlab{}.
\newblock \showarticletitle{Microsecond-scale Preemption for Concurrent $\{$GPU-accelerated$\}$$\{$DNN$\}$ Inferences}. In \bibinfo{booktitle}{\emph{16th USENIX Symposium on Operating Systems Design and Implementation (OSDI 22)}}. \bibinfo{pages}{539--558}.
\newblock


\bibitem[He et~al\mbox{.}(2016)]%
        {he2016deep}
\bibfield{author}{\bibinfo{person}{Kaiming He}, \bibinfo{person}{Xiangyu Zhang}, \bibinfo{person}{Shaoqing Ren}, {and} \bibinfo{person}{Jian Sun}.} \bibinfo{year}{2016}\natexlab{}.
\newblock \showarticletitle{Deep residual learning for image recognition}. In \bibinfo{booktitle}{\emph{Proceedings of the IEEE conference on computer vision and pattern recognition}}. \bibinfo{pages}{770--778}.
\newblock


\bibitem[Huang et~al\mbox{.}(2020)]%
        {huang2020swapadvisor}
\bibfield{author}{\bibinfo{person}{Chien-Chin Huang}, \bibinfo{person}{Gu Jin}, {and} \bibinfo{person}{Jinyang Li}.} \bibinfo{year}{2020}\natexlab{}.
\newblock \showarticletitle{Swapadvisor: Pushing deep learning beyond the gpu memory limit via smart swapping}. In \bibinfo{booktitle}{\emph{Proceedings of the Twenty-Fifth International Conference on Architectural Support for Programming Languages and Operating Systems}}. \bibinfo{pages}{1341--1355}.
\newblock


\bibitem[Jain et~al\mbox{.}(2020)]%
        {jain2020checkmate}
\bibfield{author}{\bibinfo{person}{Paras Jain}, \bibinfo{person}{Ajay Jain}, \bibinfo{person}{Aniruddha Nrusimha}, \bibinfo{person}{Amir Gholami}, \bibinfo{person}{Pieter Abbeel}, \bibinfo{person}{Joseph Gonzalez}, \bibinfo{person}{Kurt Keutzer}, {and} \bibinfo{person}{Ion Stoica}.} \bibinfo{year}{2020}\natexlab{}.
\newblock \showarticletitle{Checkmate: Breaking the memory wall with optimal tensor rematerialization}.
\newblock \bibinfo{journal}{\emph{Proceedings of Machine Learning and Systems}}  \bibinfo{volume}{2} (\bibinfo{year}{2020}), \bibinfo{pages}{497--511}.
\newblock


\bibitem[Kilburn et~al\mbox{.}(1962)]%
        {kilburn1962one}
\bibfield{author}{\bibinfo{person}{Tom Kilburn}, \bibinfo{person}{David~BG Edwards}, \bibinfo{person}{Michael~J Lanigan}, {and} \bibinfo{person}{Frank~H Sumner}.} \bibinfo{year}{1962}\natexlab{}.
\newblock \showarticletitle{One-level storage system}.
\newblock \bibinfo{journal}{\emph{IRE Transactions on Electronic Computers}} \bibinfo{number}{2} (\bibinfo{year}{1962}), \bibinfo{pages}{223--235}.
\newblock


\bibitem[Lester et~al\mbox{.}(2021)]%
        {lester2021power}
\bibfield{author}{\bibinfo{person}{Brian Lester}, \bibinfo{person}{Rami Al-Rfou}, {and} \bibinfo{person}{Noah Constant}.} \bibinfo{year}{2021}\natexlab{}.
\newblock \showarticletitle{The power of scale for parameter-efficient prompt tuning}.
\newblock \bibinfo{journal}{\emph{arXiv preprint arXiv:2104.08691}} (\bibinfo{year}{2021}).
\newblock


\bibitem[Li and Liang(2021)]%
        {li2021prefix}
\bibfield{author}{\bibinfo{person}{Xiang~Lisa Li} {and} \bibinfo{person}{Percy Liang}.} \bibinfo{year}{2021}\natexlab{}.
\newblock \showarticletitle{Prefix-tuning: Optimizing continuous prompts for generation}.
\newblock \bibinfo{journal}{\emph{arXiv preprint arXiv:2101.00190}} (\bibinfo{year}{2021}).
\newblock


\bibitem[Li et~al\mbox{.}(2023)]%
        {li2023alpaserve}
\bibfield{author}{\bibinfo{person}{Zhuohan Li}, \bibinfo{person}{Lianmin Zheng}, \bibinfo{person}{Yinmin Zhong}, \bibinfo{person}{Vincent Liu}, \bibinfo{person}{Ying Sheng}, \bibinfo{person}{Xin Jin}, \bibinfo{person}{Yanping Huang}, \bibinfo{person}{Zhifeng Chen}, \bibinfo{person}{Hao Zhang}, \bibinfo{person}{Joseph~E Gonzalez}, {et~al\mbox{.}}} \bibinfo{year}{2023}\natexlab{}.
\newblock \showarticletitle{AlpaServe: Statistical Multiplexing with Model Parallelism for Deep Learning Serving}.
\newblock \bibinfo{journal}{\emph{arXiv preprint arXiv:2302.11665}} (\bibinfo{year}{2023}).
\newblock


\bibitem[Ma et~al\mbox{.}(2020)]%
        {ma2020rammer}
\bibfield{author}{\bibinfo{person}{Lingxiao Ma}, \bibinfo{person}{Zhiqiang Xie}, \bibinfo{person}{Zhi Yang}, \bibinfo{person}{Jilong Xue}, \bibinfo{person}{Youshan Miao}, \bibinfo{person}{Wei Cui}, \bibinfo{person}{Wenxiang Hu}, \bibinfo{person}{Fan Yang}, \bibinfo{person}{Lintao Zhang}, {and} \bibinfo{person}{Lidong Zhou}.} \bibinfo{year}{2020}\natexlab{}.
\newblock \showarticletitle{Rammer: Enabling holistic deep learning compiler optimizations with rtasks}. In \bibinfo{booktitle}{\emph{Proceedings of the 14th USENIX Conference on Operating Systems Design and Implementation}}. \bibinfo{pages}{881--897}.
\newblock


\bibitem[NVIDIA({[n.\,d.]})]%
        {nvidiatriton}
\bibfield{author}{\bibinfo{person}{NVIDIA}.} \bibinfo{year}{[n.\,d.]}\natexlab{}.
\newblock \bibinfo{title}{Triton Inference Server}.
\newblock \bibinfo{howpublished}{\url{https://developer.nvidia.com/nvidia-triton-inference-server}}.
\newblock


\bibitem[NVIDIA(2023a)]%
        {nvidiaft}
\bibfield{author}{\bibinfo{person}{NVIDIA}.} \bibinfo{year}{2023}\natexlab{a}.
\newblock \bibinfo{title}{FasterTransformer}.
\newblock \bibinfo{howpublished}{\url{https://github.com/NVIDIA/FasterTransformer}}.
\newblock


\bibitem[NVIDIA(2023b)]%
        {nccl}
\bibfield{author}{\bibinfo{person}{NVIDIA}.} \bibinfo{year}{2023}\natexlab{b}.
\newblock \bibinfo{title}{NCCL: The NVIDIA Collective Communication Library}.
\newblock \bibinfo{howpublished}{\url{https://developer.nvidia.com/nccl}}.
\newblock


\bibitem[Olston et~al\mbox{.}(2017)]%
        {olston2017tensorflow}
\bibfield{author}{\bibinfo{person}{Christopher Olston}, \bibinfo{person}{Noah Fiedel}, \bibinfo{person}{Kiril Gorovoy}, \bibinfo{person}{Jeremiah Harmsen}, \bibinfo{person}{Li Lao}, \bibinfo{person}{Fangwei Li}, \bibinfo{person}{Vinu Rajashekhar}, \bibinfo{person}{Sukriti Ramesh}, {and} \bibinfo{person}{Jordan Soyke}.} \bibinfo{year}{2017}\natexlab{}.
\newblock \showarticletitle{Tensorflow-serving: Flexible, high-performance ml serving}.
\newblock \bibinfo{journal}{\emph{arXiv preprint arXiv:1712.06139}} (\bibinfo{year}{2017}).
\newblock


\bibitem[OpenAI(2020)]%
        {openaiapi}
\bibfield{author}{\bibinfo{person}{OpenAI}.} \bibinfo{year}{2020}\natexlab{}.
\newblock
\newblock
\urldef\tempurl%
\url{https://openai.com/blog/openai-api}
\showURL{%
\tempurl}


\bibitem[OpenAI(2022)]%
        {chatgpt}
\bibfield{author}{\bibinfo{person}{OpenAI}.} \bibinfo{year}{2022}\natexlab{}.
\newblock
\newblock
\urldef\tempurl%
\url{https://openai.com/blog/chatgpt}
\showURL{%
\tempurl}


\bibitem[OpenAI(2023a)]%
        {chatgptuserprompt}
\bibfield{author}{\bibinfo{person}{OpenAI}.} \bibinfo{year}{2023}\natexlab{a}.
\newblock
\newblock
\urldef\tempurl%
\url{https://openai.com/blog/custom-instructions-for-chatgpt}
\showURL{%
\tempurl}


\bibitem[OpenAI(2023b)]%
        {openai2023gpt4}
\bibfield{author}{\bibinfo{person}{OpenAI}.} \bibinfo{year}{2023}\natexlab{b}.
\newblock \bibinfo{title}{GPT-4 Technical Report}.
\newblock
\newblock
\showeprint[arxiv]{2303.08774}~[cs.CL]


\bibitem[ORG(2023)]%
        {lmsysweek8}
\bibfield{author}{\bibinfo{person}{LMSYS ORG}.} \bibinfo{year}{2023}\natexlab{}.
\newblock \bibinfo{title}{Chatbot Arena Leaderboard Week 8: Introducing MT-Bench and Vicuna-33B}.
\newblock \bibinfo{howpublished}{https://lmsys.org/blog/2023-06-22-leaderboard/}.
\newblock


\bibitem[Paszke et~al\mbox{.}(2019)]%
        {paszke2019pytorch}
\bibfield{author}{\bibinfo{person}{Adam Paszke}, \bibinfo{person}{Sam Gross}, \bibinfo{person}{Francisco Massa}, \bibinfo{person}{Adam Lerer}, \bibinfo{person}{James Bradbury}, \bibinfo{person}{Gregory Chanan}, \bibinfo{person}{Trevor Killeen}, \bibinfo{person}{Zeming Lin}, \bibinfo{person}{Natalia Gimelshein}, \bibinfo{person}{Luca Antiga}, {et~al\mbox{.}}} \bibinfo{year}{2019}\natexlab{}.
\newblock \showarticletitle{Pytorch: An imperative style, high-performance deep learning library}.
\newblock \bibinfo{journal}{\emph{Advances in neural information processing systems}}  \bibinfo{volume}{32} (\bibinfo{year}{2019}).
\newblock


\bibitem[Patil et~al\mbox{.}(2022)]%
        {patil2022poet}
\bibfield{author}{\bibinfo{person}{Shishir~G Patil}, \bibinfo{person}{Paras Jain}, \bibinfo{person}{Prabal Dutta}, \bibinfo{person}{Ion Stoica}, {and} \bibinfo{person}{Joseph Gonzalez}.} \bibinfo{year}{2022}\natexlab{}.
\newblock \showarticletitle{POET: Training Neural Networks on Tiny Devices with Integrated Rematerialization and Paging}. In \bibinfo{booktitle}{\emph{International Conference on Machine Learning}}. PMLR, \bibinfo{pages}{17573--17583}.
\newblock


\bibitem[Pope et~al\mbox{.}(2022)]%
        {pope2022efficiently}
\bibfield{author}{\bibinfo{person}{Reiner Pope}, \bibinfo{person}{Sholto Douglas}, \bibinfo{person}{Aakanksha Chowdhery}, \bibinfo{person}{Jacob Devlin}, \bibinfo{person}{James Bradbury}, \bibinfo{person}{Anselm Levskaya}, \bibinfo{person}{Jonathan Heek}, \bibinfo{person}{Kefan Xiao}, \bibinfo{person}{Shivani Agrawal}, {and} \bibinfo{person}{Jeff Dean}.} \bibinfo{year}{2022}\natexlab{}.
\newblock \showarticletitle{Efficiently Scaling Transformer Inference}.
\newblock \bibinfo{journal}{\emph{arXiv preprint arXiv:2211.05102}} (\bibinfo{year}{2022}).
\newblock


\bibitem[Ren et~al\mbox{.}(2021)]%
        {ren2021zero}
\bibfield{author}{\bibinfo{person}{Jie Ren}, \bibinfo{person}{Samyam Rajbhandari}, \bibinfo{person}{Reza~Yazdani Aminabadi}, \bibinfo{person}{Olatunji Ruwase}, \bibinfo{person}{Shuangyan Yang}, \bibinfo{person}{Minjia Zhang}, \bibinfo{person}{Dong Li}, {and} \bibinfo{person}{Yuxiong He}.} \bibinfo{year}{2021}\natexlab{}.
\newblock \showarticletitle{ZeRO-Offload: Democratizing Billion-Scale Model Training.}. In \bibinfo{booktitle}{\emph{USENIX Annual Technical Conference}}. \bibinfo{pages}{551--564}.
\newblock


\bibitem[Reuters(2023)]%
        {chat-cost}
\bibfield{author}{\bibinfo{person}{Reuters}.} \bibinfo{year}{2023}\natexlab{}.
\newblock
\newblock
\urldef\tempurl%
\url{https://www.reuters.com/technology/tech-giants-ai-like-bing-bard-poses-billion-dollar-search-problem-2023-02-22/}
\showURL{%
\tempurl}


\bibitem[Services(2023)]%
        {amazonbedrock}
\bibfield{author}{\bibinfo{person}{Amazon~Web Services}.} \bibinfo{year}{2023}\natexlab{}.
\newblock
\newblock
\urldef\tempurl%
\url{https://aws.amazon.com/bedrock/}
\showURL{%
\tempurl}


\bibitem[Shen et~al\mbox{.}(2019)]%
        {shen2019nexus}
\bibfield{author}{\bibinfo{person}{Haichen Shen}, \bibinfo{person}{Lequn Chen}, \bibinfo{person}{Yuchen Jin}, \bibinfo{person}{Liangyu Zhao}, \bibinfo{person}{Bingyu Kong}, \bibinfo{person}{Matthai Philipose}, \bibinfo{person}{Arvind Krishnamurthy}, {and} \bibinfo{person}{Ravi Sundaram}.} \bibinfo{year}{2019}\natexlab{}.
\newblock \showarticletitle{Nexus: A GPU cluster engine for accelerating DNN-based video analysis}. In \bibinfo{booktitle}{\emph{Proceedings of the 27th ACM Symposium on Operating Systems Principles}}. \bibinfo{pages}{322--337}.
\newblock


\bibitem[Sheng et~al\mbox{.}(2023)]%
        {sheng2023high}
\bibfield{author}{\bibinfo{person}{Ying Sheng}, \bibinfo{person}{Lianmin Zheng}, \bibinfo{person}{Binhang Yuan}, \bibinfo{person}{Zhuohan Li}, \bibinfo{person}{Max Ryabinin}, \bibinfo{person}{Daniel~Y Fu}, \bibinfo{person}{Zhiqiang Xie}, \bibinfo{person}{Beidi Chen}, \bibinfo{person}{Clark Barrett}, \bibinfo{person}{Joseph~E Gonzalez}, {et~al\mbox{.}}} \bibinfo{year}{2023}\natexlab{}.
\newblock \showarticletitle{High-throughput Generative Inference of Large Language Models with a Single GPU}.
\newblock \bibinfo{journal}{\emph{arXiv preprint arXiv:2303.06865}} (\bibinfo{year}{2023}).
\newblock


\bibitem[Shoeybi et~al\mbox{.}(2019)]%
        {shoeybi2019megatron}
\bibfield{author}{\bibinfo{person}{Mohammad Shoeybi}, \bibinfo{person}{Mostofa Patwary}, \bibinfo{person}{Raul Puri}, \bibinfo{person}{Patrick LeGresley}, \bibinfo{person}{Jared Casper}, {and} \bibinfo{person}{Bryan Catanzaro}.} \bibinfo{year}{2019}\natexlab{}.
\newblock \showarticletitle{Megatron-lm: Training multi-billion parameter language models using model parallelism}.
\newblock \bibinfo{journal}{\emph{arXiv preprint arXiv:1909.08053}} (\bibinfo{year}{2019}).
\newblock


\bibitem[Steiner et~al\mbox{.}(2022)]%
        {steiner2022olla}
\bibfield{author}{\bibinfo{person}{Benoit Steiner}, \bibinfo{person}{Mostafa Elhoushi}, \bibinfo{person}{Jacob Kahn}, {and} \bibinfo{person}{James Hegarty}.} \bibinfo{year}{2022}\natexlab{}.
\newblock \showarticletitle{OLLA: Optimizing the Lifetime and Location of Arrays to Reduce the Memory Usage of Neural Networks}.
\newblock  (\bibinfo{year}{2022}).
\newblock
\urldef\tempurl%
\url{https://doi.org/10.48550/arXiv.2210.12924}
\showDOI{\tempurl}


\bibitem[Sutskever et~al\mbox{.}(2014)]%
        {sutskever2014sequence}
\bibfield{author}{\bibinfo{person}{Ilya Sutskever}, \bibinfo{person}{Oriol Vinyals}, {and} \bibinfo{person}{Quoc~V Le}.} \bibinfo{year}{2014}\natexlab{}.
\newblock \showarticletitle{Sequence to sequence learning with neural networks}.
\newblock \bibinfo{journal}{\emph{Advances in neural information processing systems}}  \bibinfo{volume}{27} (\bibinfo{year}{2014}).
\newblock


\bibitem[Taori et~al\mbox{.}(2023)]%
        {alpaca}
\bibfield{author}{\bibinfo{person}{Rohan Taori}, \bibinfo{person}{Ishaan Gulrajani}, \bibinfo{person}{Tianyi Zhang}, \bibinfo{person}{Yann Dubois}, \bibinfo{person}{Xuechen Li}, \bibinfo{person}{Carlos Guestrin}, \bibinfo{person}{Percy Liang}, {and} \bibinfo{person}{Tatsunori~B. Hashimoto}.} \bibinfo{year}{2023}\natexlab{}.
\newblock \bibinfo{title}{Stanford Alpaca: An Instruction-following LLaMA model}.
\newblock \bibinfo{howpublished}{\url{https://github.com/tatsu-lab/stanford_alpaca}}.
\newblock


\bibitem[Team(2023)]%
        {sharegpt}
\bibfield{author}{\bibinfo{person}{ShareGPT Team}.} \bibinfo{year}{2023}\natexlab{}.
\newblock
\newblock
\urldef\tempurl%
\url{https://sharegpt.com/}
\showURL{%
\tempurl}


\bibitem[Touvron et~al\mbox{.}(2023)]%
        {touvron2023llama}
\bibfield{author}{\bibinfo{person}{Hugo Touvron}, \bibinfo{person}{Thibaut Lavril}, \bibinfo{person}{Gautier Izacard}, \bibinfo{person}{Xavier Martinet}, \bibinfo{person}{Marie-Anne Lachaux}, \bibinfo{person}{Timoth{\'e}e Lacroix}, \bibinfo{person}{Baptiste Rozi{\`e}re}, \bibinfo{person}{Naman Goyal}, \bibinfo{person}{Eric Hambro}, \bibinfo{person}{Faisal Azhar}, {et~al\mbox{.}}} \bibinfo{year}{2023}\natexlab{}.
\newblock \showarticletitle{Llama: Open and efficient foundation language models}.
\newblock \bibinfo{journal}{\emph{arXiv preprint arXiv:2302.13971}} (\bibinfo{year}{2023}).
\newblock


\bibitem[Vaswani et~al\mbox{.}(2017)]%
        {vaswani2017attention}
\bibfield{author}{\bibinfo{person}{Ashish Vaswani}, \bibinfo{person}{Noam Shazeer}, \bibinfo{person}{Niki Parmar}, \bibinfo{person}{Jakob Uszkoreit}, \bibinfo{person}{Llion Jones}, \bibinfo{person}{Aidan~N Gomez}, \bibinfo{person}{{\L}ukasz Kaiser}, {and} \bibinfo{person}{Illia Polosukhin}.} \bibinfo{year}{2017}\natexlab{}.
\newblock \showarticletitle{Attention is all you need}.
\newblock \bibinfo{journal}{\emph{Advances in neural information processing systems}}  \bibinfo{volume}{30} (\bibinfo{year}{2017}).
\newblock


\bibitem[Wang et~al\mbox{.}(2022b)]%
        {wang2022pacman}
\bibfield{author}{\bibinfo{person}{Jing Wang}, \bibinfo{person}{Youyou Lu}, \bibinfo{person}{Qing Wang}, \bibinfo{person}{Minhui Xie}, \bibinfo{person}{Keji Huang}, {and} \bibinfo{person}{Jiwu Shu}.} \bibinfo{year}{2022}\natexlab{b}.
\newblock \showarticletitle{Pacman: An Efficient Compaction Approach for $\{$Log-Structured$\}$$\{$Key-Value$\}$ Store on Persistent Memory}. In \bibinfo{booktitle}{\emph{2022 USENIX Annual Technical Conference (USENIX ATC 22)}}. \bibinfo{pages}{773--788}.
\newblock


\bibitem[Wang et~al\mbox{.}(2018)]%
        {wang2018superneurons}
\bibfield{author}{\bibinfo{person}{Linnan Wang}, \bibinfo{person}{Jinmian Ye}, \bibinfo{person}{Yiyang Zhao}, \bibinfo{person}{Wei Wu}, \bibinfo{person}{Ang Li}, \bibinfo{person}{Shuaiwen~Leon Song}, \bibinfo{person}{Zenglin Xu}, {and} \bibinfo{person}{Tim Kraska}.} \bibinfo{year}{2018}\natexlab{}.
\newblock \showarticletitle{Superneurons: Dynamic GPU memory management for training deep neural networks}. In \bibinfo{booktitle}{\emph{Proceedings of the 23rd ACM SIGPLAN symposium on principles and practice of parallel programming}}. \bibinfo{pages}{41--53}.
\newblock


\bibitem[Wang et~al\mbox{.}(2021)]%
        {wang2021lightseq}
\bibfield{author}{\bibinfo{person}{Xiaohui Wang}, \bibinfo{person}{Ying Xiong}, \bibinfo{person}{Yang Wei}, \bibinfo{person}{Mingxuan Wang}, {and} \bibinfo{person}{Lei Li}.} \bibinfo{year}{2021}\natexlab{}.
\newblock \showarticletitle{LightSeq: A High Performance Inference Library for Transformers}. In \bibinfo{booktitle}{\emph{Proceedings of the 2021 Conference of the North American Chapter of the Association for Computational Linguistics: Human Language Technologies: Industry Papers}}. \bibinfo{pages}{113--120}.
\newblock


\bibitem[Wang et~al\mbox{.}(2022a)]%
        {wang2022self}
\bibfield{author}{\bibinfo{person}{Yizhong Wang}, \bibinfo{person}{Yeganeh Kordi}, \bibinfo{person}{Swaroop Mishra}, \bibinfo{person}{Alisa Liu}, \bibinfo{person}{Noah~A Smith}, \bibinfo{person}{Daniel Khashabi}, {and} \bibinfo{person}{Hannaneh Hajishirzi}.} \bibinfo{year}{2022}\natexlab{a}.
\newblock \showarticletitle{Self-Instruct: Aligning Language Model with Self Generated Instructions}.
\newblock \bibinfo{journal}{\emph{arXiv preprint arXiv:2212.10560}} (\bibinfo{year}{2022}).
\newblock


\bibitem[Wolf et~al\mbox{.}(2020)]%
        {wolf2020transformers}
\bibfield{author}{\bibinfo{person}{Thomas Wolf}, \bibinfo{person}{Lysandre Debut}, \bibinfo{person}{Victor Sanh}, \bibinfo{person}{Julien Chaumond}, \bibinfo{person}{Clement Delangue}, \bibinfo{person}{Anthony Moi}, \bibinfo{person}{Pierric Cistac}, \bibinfo{person}{Tim Rault}, \bibinfo{person}{R{\'e}mi Louf}, \bibinfo{person}{Morgan Funtowicz}, {et~al\mbox{.}}} \bibinfo{year}{2020}\natexlab{}.
\newblock \showarticletitle{Transformers: State-of-the-art natural language processing}. In \bibinfo{booktitle}{\emph{Proceedings of the 2020 conference on empirical methods in natural language processing: system demonstrations}}. \bibinfo{pages}{38--45}.
\newblock


\bibitem[Wu et~al\mbox{.}(2016)]%
        {wu2016google}
\bibfield{author}{\bibinfo{person}{Yonghui Wu}, \bibinfo{person}{Mike Schuster}, \bibinfo{person}{Zhifeng Chen}, \bibinfo{person}{Quoc~V Le}, \bibinfo{person}{Mohammad Norouzi}, \bibinfo{person}{Wolfgang Macherey}, \bibinfo{person}{Maxim Krikun}, \bibinfo{person}{Yuan Cao}, \bibinfo{person}{Qin Gao}, \bibinfo{person}{Klaus Macherey}, {et~al\mbox{.}}} \bibinfo{year}{2016}\natexlab{}.
\newblock \showarticletitle{Google's neural machine translation system: Bridging the gap between human and machine translation}.
\newblock \bibinfo{journal}{\emph{arXiv preprint arXiv:1609.08144}} (\bibinfo{year}{2016}).
\newblock


\bibitem[Yu et~al\mbox{.}(2022)]%
        {yu2022orca}
\bibfield{author}{\bibinfo{person}{Gyeong-In Yu}, \bibinfo{person}{Joo~Seong Jeong}, \bibinfo{person}{Geon-Woo Kim}, \bibinfo{person}{Soojeong Kim}, {and} \bibinfo{person}{Byung-Gon Chun}.} \bibinfo{year}{2022}\natexlab{}.
\newblock \showarticletitle{Orca: A Distributed Serving System for $\{$Transformer-Based$\}$ Generative Models}. In \bibinfo{booktitle}{\emph{16th USENIX Symposium on Operating Systems Design and Implementation (OSDI 22)}}. \bibinfo{pages}{521--538}.
\newblock


\bibitem[Zhang et~al\mbox{.}(2023)]%
        {zhang2023shepherd}
\bibfield{author}{\bibinfo{person}{Hong Zhang}, \bibinfo{person}{Yupeng Tang}, \bibinfo{person}{Anurag Khandelwal}, {and} \bibinfo{person}{Ion Stoica}.} \bibinfo{year}{2023}\natexlab{}.
\newblock \showarticletitle{{SHEPHERD}: Serving {DNNs} in the Wild}. In \bibinfo{booktitle}{\emph{20th USENIX Symposium on Networked Systems Design and Implementation (NSDI 23)}}. \bibinfo{publisher}{USENIX Association}, \bibinfo{address}{Boston, MA}, \bibinfo{pages}{787--808}.
\newblock
\showISBNx{978-1-939133-33-5}
\urldef\tempurl%
\url{https://www.usenix.org/conference/nsdi23/presentation/zhang-hong}
\showURL{%
\tempurl}


\bibitem[Zhang et~al\mbox{.}(2022)]%
        {zhang2022opt}
\bibfield{author}{\bibinfo{person}{Susan Zhang}, \bibinfo{person}{Stephen Roller}, \bibinfo{person}{Naman Goyal}, \bibinfo{person}{Mikel Artetxe}, \bibinfo{person}{Moya Chen}, \bibinfo{person}{Shuohui Chen}, \bibinfo{person}{Christopher Dewan}, \bibinfo{person}{Mona Diab}, \bibinfo{person}{Xian Li}, \bibinfo{person}{Xi~Victoria Lin}, {et~al\mbox{.}}} \bibinfo{year}{2022}\natexlab{}.
\newblock \showarticletitle{Opt: Open pre-trained transformer language models}.
\newblock \bibinfo{journal}{\emph{arXiv preprint arXiv:2205.01068}} (\bibinfo{year}{2022}).
\newblock


\bibitem[Zheng et~al\mbox{.}(2022)]%
        {zheng2022alpa}
\bibfield{author}{\bibinfo{person}{Lianmin Zheng}, \bibinfo{person}{Zhuohan Li}, \bibinfo{person}{Hao Zhang}, \bibinfo{person}{Yonghao Zhuang}, \bibinfo{person}{Zhifeng Chen}, \bibinfo{person}{Yanping Huang}, \bibinfo{person}{Yida Wang}, \bibinfo{person}{Yuanzhong Xu}, \bibinfo{person}{Danyang Zhuo}, \bibinfo{person}{Eric~P Xing}, {et~al\mbox{.}}} \bibinfo{year}{2022}\natexlab{}.
\newblock \showarticletitle{Alpa: Automating Inter-and Intra-Operator Parallelism for Distributed Deep Learning}. In \bibinfo{booktitle}{\emph{16th USENIX Symposium on Operating Systems Design and Implementation (OSDI 22)}}. \bibinfo{pages}{559--578}.
\newblock


\bibitem[Zhou et~al\mbox{.}(2022)]%
        {zhou2022pets}
\bibfield{author}{\bibinfo{person}{Zhe Zhou}, \bibinfo{person}{Xuechao Wei}, \bibinfo{person}{Jiejing Zhang}, {and} \bibinfo{person}{Guangyu Sun}.} \bibinfo{year}{2022}\natexlab{}.
\newblock \showarticletitle{PetS: A Unified Framework for Parameter-Efficient Transformers Serving}. In \bibinfo{booktitle}{\emph{2022 USENIX Annual Technical Conference (USENIX ATC 22)}}. \bibinfo{pages}{489--504}.
\newblock


\end{thebibliography}

\end{document}